\documentclass[]{bytedance_seed}

\usepackage[toc,page,header]{appendix}

\usepackage{minitoc}

\usepackage{amsmath,amsfonts,bm}

\def\eqref#1{equation~\ref{#1}}

\def\1{\bm{1}}

\DeclareMathAlphabet{\mathsfit}{\encodingdefault}{\sfdefault}{m}{sl}
\SetMathAlphabet{\mathsfit}{bold}{\encodingdefault}{\sfdefault}{bx}{n}

\usepackage{wrapfig}
\usepackage{multirow}
\usepackage{enumitem}
\usepackage{adjustbox}
\usepackage{tabularx}
\usepackage{listings}
\usepackage{soul}
\usepackage{threeparttable}
\usepackage{tcolorbox}
\usepackage{array}
\usepackage{parskip}

\lstnewenvironment{prompt}[1]
{
    \lstset{
      caption=#1,
      captionpos=b,
      basicstyle=\small,
      columns=fullflexible,
      frame=single,
      breaklines=true,
      breakindent=0pt,
      backgroundcolor=\color{gray!10}
    }
}{}

\newcommand{\ours}{\textbf{MGA}}

\title{Reformulation for Pretraining Data Augmentation}
\author[1]{Xintong Hao}
\author[1,2]{Ruijie Zhu}
\author[1]{Ge Zhang}
\author[1]{Ke Shen}
\author[1]{Chenggang Li}

\affiliation[1]{ByteDance Seed}
\affiliation[2]{University of California, Santa Cruz}

\abstract{

Despite the impressive capabilities of large language models across various tasks, their continued scaling is severely hampered not only by data scarcity but also by the performance degradation associated with excessive data repetition during training. 
To overcome this critical bottleneck, we propose the Massive Genre-Audience~(\textbf{MGA}) reformulation method, a lightweight and scalable data augmentation technique inspired by synthetic data methodologies. MGA systematically reformulates existing corpora into diverse, contextually-rich variations to mitigate the negative effects of repetition, and we introduce this approach along with the resulting \textbf{770 billion token MGACorpus} in this work.
We experimentally validate its core benefit by demonstrating superior performance against data repetition and upsampling in scaling scenarios (up to 13B parameters).
Furthermore, comprehensive analysis investigates the role of prompt engineering in generation quality and reveals nuances in evaluating model capabilities using standard loss metrics. 
Our work shows that \textbf{MGA} provides a reliable pathway to substantially augment training datasets, effectively alleviating repetition bottlenecks and enabling more efficient scaling of large language models.

}

\date{\today}

\checkdata[Project Page]{\url{https://huggingface.co/datasets/ByteDance-Seed/mga-fineweb-edu}}

\begin{document}
\maketitle


\section{Introduction}
\label{sec}
\vspace{-0.5em}

The remarkable success of Large Language Models (LLMs) heavily relies on the scale of model parameters and training data~\citep{kaplan2020scaling,hoffmann2022empirical}. Scaling laws demonstrate that improvements in model performance are increasingly dependent on data quantity and quality. However, the growth rate of available natural language corpora significantly lags behind the increasing demand for training data~\citep{villalobos2022will}. In traditional deep learning, data repetition has been a standard approach—training models for over 1,000 epochs on ImageNet is common and continues to yield improvements. Yet, in the pre-training stage of LLMs, excessive data repetition can degrade model performance and stability, creating a significant barrier to continued scaling efforts, particularly for the largest models. This raises a critical question: how can we fully utilize the potential of existing data in data-constrained situations?

Data augmentation has been widely employed to address similar challenges in traditional machine learning. However, conventional augmentation methods have proven ineffective for LLMs. One emerging approach involves leveraging LLMs themselves to synthesize high-quality training data~\citep{su2024nemotron,abdin2024phi}. Data synthesis could theoretically generate limitless diverse training material, enabling dataset expansion without the negative consequences associated with excessive repetition.

However, prevailing data synthesis methods face significant hurdles. Many depend on large-scale models for generation, such as 12B dense models~\citep{su2024nemotron} or those with GPT-4-level capabilities~\citep{abdin2024phi}, to ensure data quality. This effectively transforms synthetic datasets into ``distillations'' from larger models rather than true data augmentations. Also, using such large models to generate additional data during the actual pre-training process is clearly impractical from a computational standpoint. Furthermore, some approaches, like Phi and Cosmopedia, require sophisticated, pre-defined seed curation systems to manage data diversity~\citep{abdin2024phi, benallal2024smollmcorpus}. This dual reliance on massive models and complex seed management introduces substantial computational bottlenecks and scalability challenges, limiting their practicality for efficient pretraining corpus expansion specifically aimed at mitigating data repetition issues.

\begin{figure*}[t]
\vspace{-2em}
\centering
\includegraphics[width=1\columnwidth]{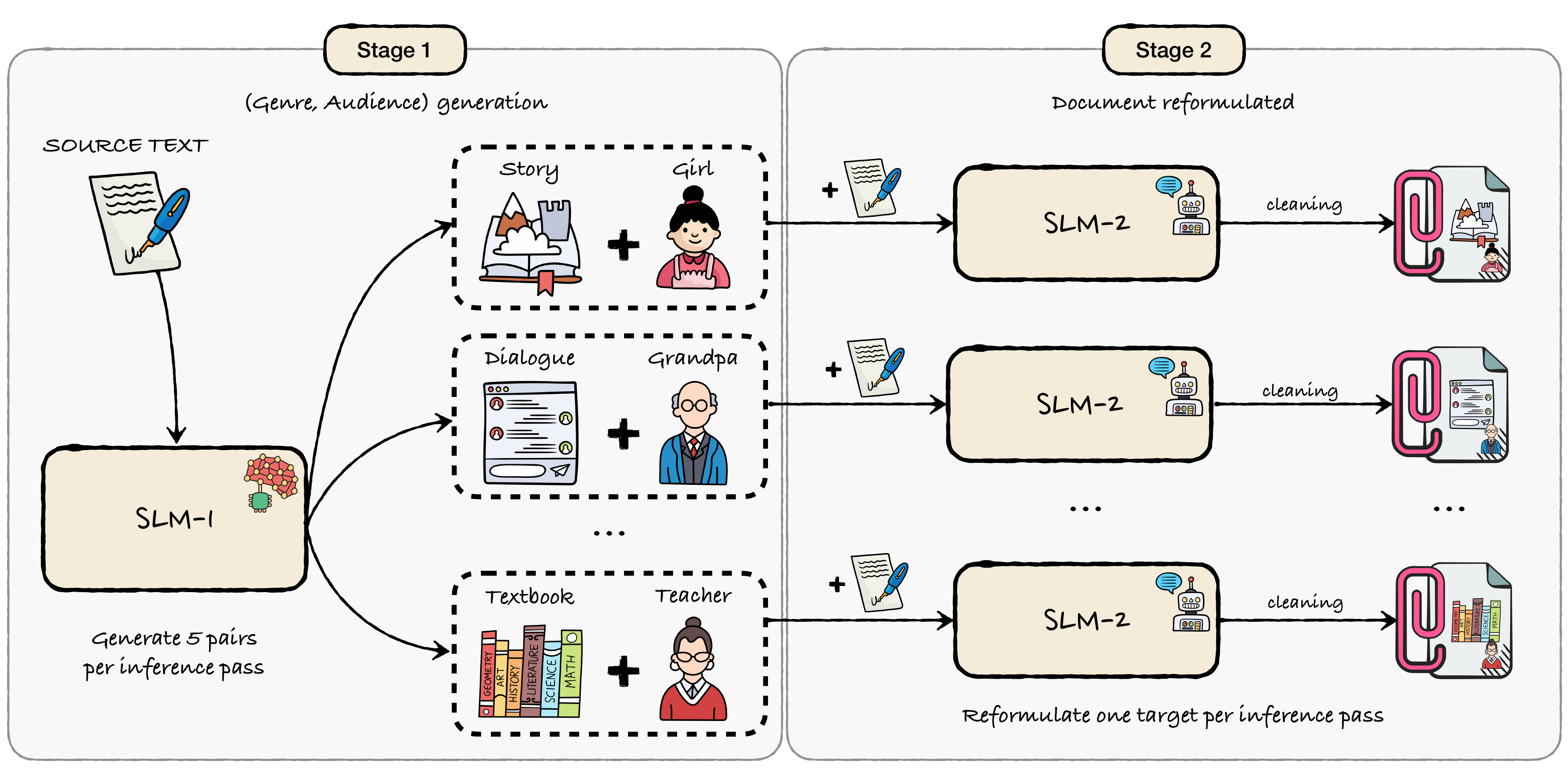}
\vspace{-2em}
\caption{Overview of {\ours} framework. Our method expands the original corpus through a two-stage synthesis process. Each document is reformulated to 5 new documents, achieving 3.9× token number expansion while maintaining diversity through massive (genre, audience) pairs.}
\label{fig:method-overview}
\vspace{-0.5em}
\end{figure*}

In this work, we propose {\ours} (Massive Genre-Audience reformulation), a more efficient approach designed to directly address the data repetition challenge. As illustrated in~\autoref{fig:method-overview}, {\ours} utilizes a comparatively lightweight 3.3B MoE model. Crucially, it avoids complex external seed systems by adaptively generating diverse genre-audience pairs directly from raw input documents. This makes the process lightweight and scalable, offering a practical way to expand datasets while minimizing detrimental repetition. Our main contributions are:

\begin{itemize}[leftmargin=1.5em]
\item We build and introduce the MGACorpus, a 770 billion token dataset based on existing high-quality text collections. We demonstrate that the MGACorpus achieves superior performance compared to the original corpus it expands upon, and also shows improved results against models trained on other synthetic datasets, underscoring the quality and effectiveness of the MGA approach. 
\item We further perform a representative evaluation of data budget scaling strategies from a data augmentation perspective, revealing that MGACorpus yields consistent improvements across various model sizes (377M/1.7B/7B/13B) compared to data repetition and upsampling methods.
\item We analyze synthetic data collapse from two key perspectives, characterizing prompt engineering's mitigating effects while revealing limitations of validation loss as a collapse detection metric, providing insights for future synthetic data optimization.
\end{itemize}
\section{Related Work}
\vspace{-0.5em}
\paragraph{\textbf{Data Curation}}
While web-crawled data contains hundreds of trillions of tokens, stringent quality filters typically remove the majority of this content.
Popular datasets like C4, Gopher, Dolma, RefinedWeb~\citep{raffel2020exploring,rae2021gopher,penedo2023refinedweb,soldaini2024dolma} use nonlearned heuristics method.
And recently FineWeb-Edu~\citep{penedo2024finewebdatasets}, DCLM~\citep{li2024datacomplm}, FineFineWeb~\citep{zhang2024finefineweb} focus on aggressive model-based and retrieval-based filtering. 
Such heavy filtering results in removal of 90\% of tokens, some researchers turn their attention to balance accuracy and data quantity~\citep{su2024nemotron}.
However, this does not alter the fact that the total amount of high-quality data remains limited.

\vspace{-0.5em}

\paragraph{\textbf{Repetition Training}}
Studies on subset repetition training have revealed that model divergence tends to occur earlier as model parameters increase~\citep{hernandez2022scaling}.
For scenarios training on entire datasets repeated, limiting to 4 epochs or fewer results in minimal efficiency degradation~\citep{muennighoff2023scaling,taylor2022galactica}.
Furthermore, \citep{xue2024repeat} shows that some regularization techniques (e.g., dropout) and leveraging MoE architecture can help efficient LLM development on a broader scale.
Overall, this topic remains understudied across different model architectures, data distributions, and repetition ratios.

\vspace{-0.5em}

\paragraph{\textbf{Synthetic Pretrain}}
Current synthetic data generation methods for language model pretraining can be primarily categorized into two approaches: seed based synthesis and raw text based rephrasing. 
The seed based method, exemplified by Phi-4~\citep{abdin2024phi} and Cosmopedia~\citep{benallal2024smollmcorpus}, employs predefined seed systems and task templates to precisely control the type and structure of generated content. 
The rephrasing method, represented by WRAP~\citep{maini2024rephrasing} and Nemotron-CC~\citep{su2024nemotron}, generates data by rephrasing web content into QA pairs and wiki-style texts, 
demonstrating significant effectiveness in processing noisy web text, though its benefits may be limited when applied to high-quality source data~\citep{pieler2024rephrasing}.
Additionally,~\citet{ge2024scalingpersonas} introduce an innovative text generation method based on billion personas, offering new insights into enhancing the diversity of synthetic data.

While existing approaches have made significant progress, they face key limitations:
seed-based methods require complex initialization systems, limiting investigation of their scaling properties,
while rephrasing-based approaches struggle to effectively augment high-quality corpus at scale.
To address these limitations, our framework {\ours} bridges this gap by leveraging and extending the inherent diversity of existing corpus.
Specifically, it adaptively generates multiple \textbf{Genre} and \textbf{Audience} seeds for each document of SmolLM-Corpus~\citep{benallal2024smollmcorpus}, enabling 3.9x token expansion while maintaining diversity and quality.
\section{Massive Genre-Audience Reformulation}
\label{sec:syntheses_pipeline}
\label{sec:implementation_details}
\vspace{-0.5em}
Preserving core knowledge while adapting content presentation for diverse audiences is the key motivation behind the MGA reformulation framework. As shown in \autoref{fig:method-overview}, the approach systematically expands original corpora through a two-stage synthesis process complemented by heuristic cleaning. Our implementation consists of three key components\footnote{Tool model details presented in Appendix~\ref{sec:appd_implementation_details}. Prompts and case studies in Appendix~\ref{sec:appd_prompt}.}: (1) A large language model serving as both LLM labeler and judger; (2) Task-specific Tool Models (Tool SLMs) applying W8A8 quantized~\citep{xiao2023smoothquant} for efficiency; and (3) A balanced quality assessment mechanism defined as Limited Consistency. In the following sections, we first introduce the concept of genre-audience pairs that drive content diversity, then describe the reformulation process and quality evaluation framework, followed by detailed prompt engineering strategies that ensure optimal balance between information preservation and content variation.

\subsection{Genre-Audience Pairs}
\vspace{-0.5em}
For pretraining corpus, there is a consensus among researchers to ensure diversity and quality.
Inspired by~\cite{maini2024rephrasing};\cite{ge2024scalingpersonas}, we expand the simple rephrasing method from only few styles to massive genre-audience pairs. 

\begin{itemize}[leftmargin=1.5em]
\item \textbf{Genre} defines the knowledge expression framework through multiple dimensions: communication purpose (e.g., education, analysis), content structure (e.g., step-by-step tutorials, analytical reports), language style, and knowledge depth control. This framework guides how information should be reconstructed while preserving core concepts.
\item \textbf{Audience} profiles combine demographic factors (age, education, profession) with knowledge background and motivational characteristics. For example, a beginner-level first-aid guide would be reformulated differently for medical students versus office workers, while maintaining essential medical accuracy.
\end{itemize}

Our framework supports N genres and M audience types, theoretically enabling N×M unique reformulation patterns.
To balance diversity and computational efficiency, we generate 5 genre-audience pairs per inference pass.
This ensures more distinct reformulations per document than typical N-epoch repetitions (e.g., N$\leq$4) often considered safe in LLM pretraining~\citep{muennighoff2023scaling}, aiming for novel augmentation while managing generation costs.

\subsection{Reformulation}
\vspace{-0.5em}
Once the genre-audience pairs are determined, the reformulation process follows a straightforward approach,
as prompt presented in Appendix~\ref{sec:appd_prompt2}.
The key factor of reformulation is how to evaluate the output text,
so we introduce the concept of \textbf{``Limited Consistency''} as criterion for quality controlling. 
This framework seeks to establish an optimal balance between textual variation and information preservation as shown in Prompt~1.

\begin{table*}[h]
\vspace{-0.5em}
\begin{prompt}{LLM judger prompt snippet.}
# Detailed Requirements
For scoring judgment, the following standards must be followed:
1. The `scoring range' is 1-5 points. You need to analyze and grasp each point mentioned in #Thought Process#, and give scores with distinction. Be strict, don't be too lenient with scoring!
2. The `Reformulated Text' is allowed to differ from the `Original Text' in writing style, expression style, and focus points! This cannot be a basis for deducting points!
3. The `Reformulated Text' is allowed to omit some information from the `Original Text'! Not all information from the `Original Text' needs to be reflected in the `Reformulated Text'!

The following situations will [NOT REDUCE] the score:
1. The `Reformulated Text' can include information points not present in the `Original Text'
2. The additional content in the `Reformulated Text' deviates significantly from the core information of the `Original Text'
3. The expression style, order, and focus points of the `Reformulated Text' differ from the `Original Text'

The following situations will [REDUCE] the score:
1. The information points in the `Reformulated Text' differ so greatly from the `Original Text' that it's not apparent it was Reformulated from the `Original Text'
2. The `Reformulated Text' lacks every information points present in the `Original Text'
\end{prompt}
\vspace{-1em}
\end{table*}

In practice, we use the proportion of samples (score $\ge$ 3) as our primary metric during both labeler LLM prompt engineering and tool model development. 
As shown in \autoref{table:tool_model_scores}, both our LLM and SLM achieve over 92\% with only a minor performance gap (-1.05\%).

\begin{table*}[h]

    \centering
    \caption{Performance comparison between SLM and LLM on reformulation quality evaluation.}
    \vspace{-0.5em}
    \begin{adjustbox}{max width=\textwidth}
    \setlength{\tabcolsep}{1mm}{
    \begin{tabular}{lcccccccc}
        \toprule
        Models & Total & 5 & 4 & 3 & 2 & 1 & Rate($\ge$ 3) & Diff \\
        \midrule
        Labeler LLM & 15,355 & 4,120 & 7,143 & 3,034 & 661 & 214 & 93.11\% & - \\
        Tool SLM & 15,355 & 3,788 & 7,124 & 3,224 & 736 & 285 & 92.06\% & -1.05\% \\
        \bottomrule
    \end{tabular}
    }
    \end{adjustbox}

\label{table:tool_model_scores}
\end{table*}

This trade-off between flexibility and fidelity is critical for maintaining reformulation quality while ensuring meaningful content adaptation. The empirical effects of different consistency levels are further explored in our ablation studies (Section~\ref{sec:ablation_pe}), where we demonstrate how models perform when deviating from this balanced approach.

\subsection{Prompt Engineering Strategies}
\label{sec:method_pe}
\vspace{-0.5em}

The `Limited Consistency' framework is pivotal in balancing `variance' (content variation) and `invariance' (information preservation) during reformulation. To understand how prompt design impacts this balance, corpus quality, and downstream model performance, this section quantitatively analyzes two distinct prompt engineering strategies through controlled experiments.

\vspace{-0.5em}

\paragraph{\textbf{Information Preservation Trade-off}} Since textual variance and information preservation are two conflicting yet equally critical objectives during reformulation, it is crucial to identify an optimal operating point for our prompts. We design two prompt variants:
(1) a strict version that enforces high fidelity to source information,
and (2) a relaxed version that allows substantial deviations while maintaining basic topical relevance, details could be found in Appendix~\ref{sec:appd_prompt2}.
Using these prompts, we collect training data from the same samples to train SLM variants,
denoted as SLM-Strict and SLM-Relaxed respectively, with performance presented in~\autoref{table:ablation_pe_scores}.

\begin{table*}[h]
    
    \centering
    \caption{Performance comparison of different SLM variants on reformulation quality metrics.}
    \vspace{-0.5em}
    \begin{adjustbox}{max width=0.9\textwidth}
    \setlength{\tabcolsep}{1mm}{
    \begin{tabular}{lcrrcrrccc}
        \toprule
        Models & Total & \multicolumn{1}{c}{5} & \multicolumn{1}{c}{4} & 3 & \multicolumn{1}{c}{2} & \multicolumn{1}{c}{1} & Rate($\leq 2$) & Rate($\geq 4$) & Rate($= 5$) \\
        \midrule
        SLM-Base   & 15,355 & 3,788 & 7,124 & 3,224 &   736 &   285 & 6.65\% & 71.06\% & 24.67\% \\
        SLM-Strict & 15,355 & 6,814 & 5,220 & 2,384 &   520 &   227 & 4.86\% & \textbf{78.37\%} & \textbf{44.38\%} \\
        SLM-Relaxed& 15,355 &   408 & 1,685 & 3,889 & 4,156 & 5,086 & \textbf{60.19\%} & 13.63\% & 2.66\% \\
        \bottomrule
    \end{tabular}
    }
    \end{adjustbox}

\label{table:ablation_pe_scores}
\end{table*}

\paragraph{\textbf{Distributional Analysis of Prompt Engineering Strategies}}

\begin{figure*}[h] \centering \vspace{-1em} \includegraphics[width=\linewidth]{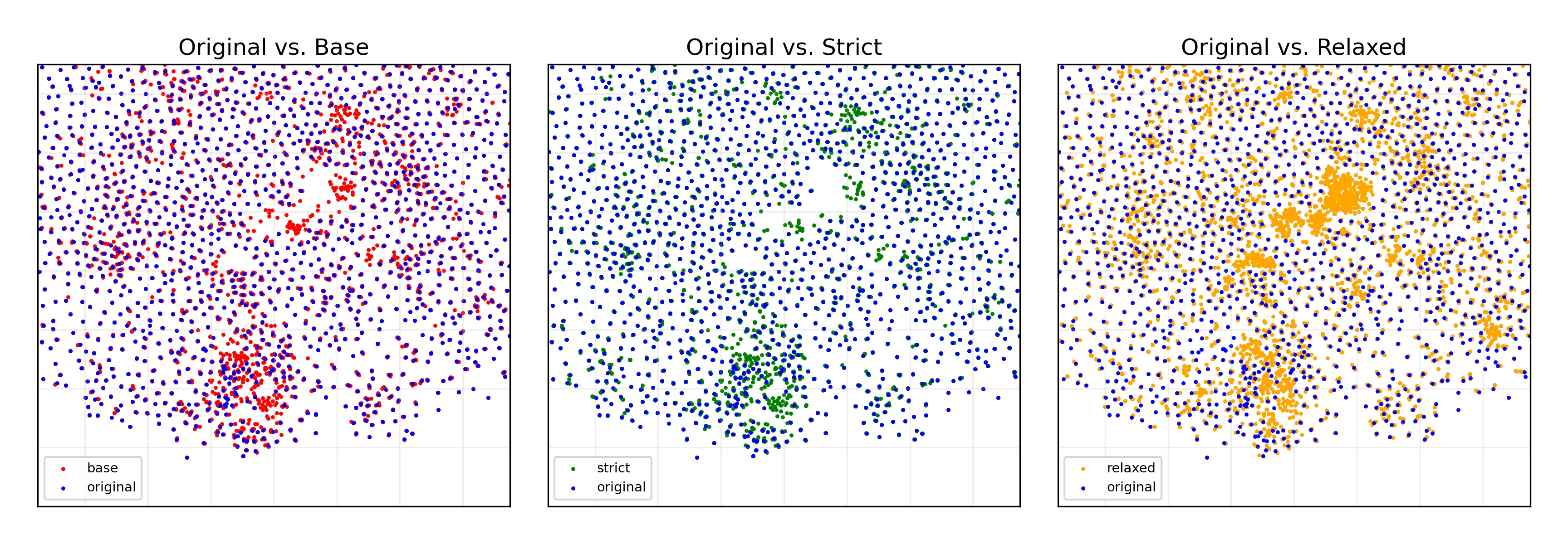} 
    \vspace{-2.5em} 
    \caption{t-SNE visualization results. Base (left) maintains a distribution that overlaps with but extends beyond the original data. Strict (middle) clusters also extend original data but indicate limited diversity compared to the Base variant. Relaxed (right) shows significant distributional shift, explaining its poor performance.} 
    \label{fig:ablation_pe_tsne} 
    \vspace{-0.5em} 
\end{figure*}

To further clarify how different prompt designs affect the resulting synthetic corpus distribution, we visualize the embeddings of documents generated by each SLM variant using t-SNE (\autoref{fig:ablation_pe_tsne}).
As illustrated, the \textbf{SLM-Base} variant produces a balanced embedding distribution, effectively expanding beyond the original data while maintaining substantial overlap. 
In contrast, the \textbf{SLM-Strict} variant demonstrates a more constrained distribution, closely adhering to the original corpus and thus limiting diversity. 
On the other hand, the \textbf{SLM-Relaxed} variant exhibits a significant distributional shift, deviating extensively from the original space, which explains its inferior performance shown in Section~\ref{sec:ablation_pe}.

These visualization results highlight the importance of carefully calibrated prompt engineering targets to achieve a desirable balance between corpus diversity and distributional coherence.
\section{Experiments}
\vspace{-0.5em}
\label{sec:experiments}
Having established our MGACorpus generation framework, we now evaluate its effectiveness through scaling experiments under data repetition scenarios in Section~\ref{sec:main_exp}. Then we present a series of experiments in Section~\ref{sec:discussions} to address the following key research questions:
\begin{itemize}[leftmargin=1.5em]

    \item \textbf{RQ1:} How effective is reformulation as a pretraining data augmentation strategy?
    \item \textbf{RQ2:} What role does reformulation diversity play in high-repetition training?
    \item \textbf{RQ3:} Why MGA reformulation benefits pretraining performance?
    
\end{itemize}
To address these research questions progressively, Section~\ref{sec:main_exp} first establishes the overall effectiveness of our MGA reformulation as a pretraining data augmentation strategy (RQ1). Subsequently, Section~\ref{sec:discussions} delves into the role of reformulation diversity in high-repetition training (RQ2) and explores the underlying reasons for MGA's benefits (RQ3).

\subsection{Setup}
\vspace{-0.5em}
\paragraph{\textbf{Datasets}}
To ensure reproducibility, we build MGACorpus based on SmolLM-Corpus~\citep{benallal2024smollmcorpus}, which contains four subsources (fineweb-edu-dedup/cosmopedia/python-edu/open-web-math), expanding fineweb-edu-dedup source from 195B tokens to 770B tokens.

\paragraph{\textbf{Models and Hyperparams}}
The architecture of pretraining model follows llama3 \citep{dubey2024llama}.
Experiments across various sizes (134M/377M/1.7B/7B/13B) are running with Warmup-Stable-Decay lr scheduler~\citep{hu2024minicpm} where 0.1\% warmup steps, 75\% stable and final 25\% decay phase.
Detailed model specifications are provided in \autoref{sec:appd_training}.

\paragraph{\textbf{Evaluation}}
We follow popular practice of \textsc{LightEval}~\citep{lighteval} and \textsc{LM-Harness}~\citep{evalharness}, evaluate on a comprehensive suite of open benchmarks
include ARC-Easy/Challenge~\citep{clark2018think}, HellaSwag~\citep{zellers2019hellaswag}, Winogrande~\citep{sakaguchi2021winogrande}, MMLU~\citep{hendrycks2020measuring}, GSM8K~\citep{cobbe2021gsm8k}, etc.
For training dynamics, we report the average of 12 benchmarks and validation losses on held-out fineweb-edu-dedup data.
And for comparison with other models, we evaluate MGACorpus aligned with Fineweb/SmolLM/Cosmopedia settings\footnote{\url{https://github.com/huggingface/cosmopedia/blob/main/evaluation}}.
While model performance is influenced by multiple factors, we list some recently SOTA models as reference, all the models are evaluated in the same environment except Llama-3.2-1B\footnote{Our access request is rejected by repo authors, so we use scores reported by SmolLM.}.

\subsection{Main Experiments}
\label{sec:main_exp}
\vspace{-0.5em}
To directly evaluate MGA's potential as a solution for data scarcity and repetition, we present a comprehensive analysis in two parts. First, we benchmark MGA's performance against recent SOTA small LMs to establish a comparative baseline. Subsequently, we investigate its behavior under data-constrained scaling scenarios, specifically situations where the training budget exceeds the available unique high-quality data, a common limitation in practical applications.

\begin{table*}[h]
    \center
    \vspace{-0.5em}
    \setlength{\fboxsep}{1pt}
    \caption{Benchmark {\ours} with SOTA small LMs. Models of similar size are grouped.
    All results are obtained through \textsc{LightEval}~\citep{lighteval}.
    Best results in each group are highlighted in \textbf{bold}, the second in \underline{underline}, and in \colorbox{green!15}{green} for that MGA wins under fair comparison.
    }
    \renewcommand{\arraystretch}{1.15}
    \begin{adjustbox}{max width=\textwidth}

    \setlength{\tabcolsep}{1mm}{
    \setlength{\fboxsep}{1.5pt}
    \begin{tabular}{@{\extracolsep{\fill}}lcccccccccccccccc}
    \toprule

      Model & \#Params. & \#Tokens & ARC(C+E) & Wino. & Hella. & MMLU & MMLU-PRO & CSQA & OpenBookQA & PIQA & TriviaQA & GSM8K & Avg. \\
      \midrule
      SmolLM2-135M & 135M & 2T & \textbf{44.12} & 51.07 & \textbf{42.03} & \textbf{31.27} & 11.06 & \underline{33.82} & 35 & \textbf{68.23} & \underline{1.91} & \textbf{1.52} & \textbf{32.00} \\
      SmolLM-135M & 135M & 600B & 42.47 & 51.54 & 41.08 & 29.93 & \underline{11.4} & 32.51 & 33.2 & \underline{68.17} & 1.08 & 0.99 & 31.24 \\
      Baseline & 134M & 600B & 41.71 & \textbf{52.41} & 40.69 & 30.03 & 11.37 & \textbf{34.32} & \underline{35.4} & 67.85 & 0.02 & 1.29 & 31.51 \\
      MGA-Expansion & 134M & 600B & \colorbox{green!15}{\underline{43.01}} & \underline{51.7} & \colorbox{green!15}{\underline{41.25}} & \colorbox{green!15}{\underline{30.1}} & \colorbox{green!15}{\textbf{11.76}} & 32.68 & \colorbox{green!15}{\textbf{36.4}} & 67.3 & \colorbox{green!15}{\textbf{2.05}} & \colorbox{green!15}{\underline{1.44}} & \colorbox{green!15}{\underline{31.77}} \\
      \midrule
      Qwen2.5-0.5B & 360M & 18T & 45.16 & \textbf{53.99} & 51.16 & 33.51 & \textbf{11.97} & 31.61 & 37.6 & 69.97 & 3.96 & \textbf{32.9} & \underline{37.18} \\
      SmolLM2-360M & 360M & 4T & \textbf{53.4} & 52.33 & \textbf{54.58} & \textbf{35.29} & 11.17 & \textbf{37.92} & 37.6 & 71.76 & \textbf{16.73} & \underline{2.96} & \textbf{37.37} \\
      SmolLM-360M & 360M & 600B & \underline{49.99} & \underline{52.96} & \underline{51.67} & 33.84 & \underline{11.42} & 34.81 & 37.6 & \underline{71.87} & 2.27 & 1.97 & 34.84 \\
      Baseline & 377M & 600B & 48.57 & 52.64 & 51.02 & 33.63 & 11.25 & 36.77 & \textbf{39} & 71 & 0.29 & 1.52 & 34.57 \\
      MGA-Expansion & 377M & 600B & \colorbox{green!15}{49.39} & 52.64 & \colorbox{green!15}{51.34} & \colorbox{green!15}{\underline{34.09}} & \colorbox{green!15}{11.35} & \colorbox{green!15}{\underline{37.1}} & \underline{38} & \colorbox{green!15}{\textbf{72.31}} & \colorbox{green!15}{\underline{7.28}} & \colorbox{green!15}{1.74} & \colorbox{green!15}{35.52} \\
      \midrule
      Qwen2.5-1.5B & 1.3B & 18T & 58.36 & \underline{58.64} & \underline{66.39} & 40.23 & 13.85 & 34.4 & 39.6 & 75.95 & 20.51 & \textbf{60.8} & \underline{46.87} \\
      SmolLM2-1.7B & 1.7B & 11T & \textbf{60.42} & \textbf{59.59} & \textbf{68.73} & \textbf{41.4} & \textbf{19.61} & \textbf{43.65} & 42.6 & \textbf{77.53} & \textbf{36.68} & \underline{29.04} & \textbf{47.93} \\
      Llama-3.2-1B & 1.2B & 9T & 49.2 & 57.8 & 61.2 & 36.63 & 11.7 & 41.2 & 38.4 & 74.8 & \underline{28.1} & 7.2 & 40.62 \\
      OLMo-1B-0724 & 1B & 3.05T & 44.71 & 56.04 & 64.38 & 32.3 & 11.8 & 33.09 & 38 & 75.24 & 13.82 & 2.43 & 37.18 \\
      SmolLM-1.7B & 1.7B & 1T & 59.95 & 54.7 & 62.83 & 39.35 & 10.92 & 38 & 42.6 & 75.9 & 13.14 & 4.62 & 40.20 \\
      Baseline & 1.7B & 1T & 59.63 & 57.38 & 65.19 & 39.4 & 12.11 & \underline{42.59} & \textbf{45.6} & 76.88 & 4.95 & 7.81 & 41.15 \\
      MGA-Expansion & 1.7B & 1T & \colorbox{green!15}{\underline{60.36}} & \colorbox{green!15}{57.46} & \colorbox{green!15}{65.52} & \colorbox{green!15}{\underline{40.79}} & \colorbox{green!15}{\underline{14.1}} & 41.11 & \underline{42.8} & \colorbox{green!15}{77.53} & \colorbox{green!15}{20.42} & \colorbox{green!15}{13.87} & \colorbox{green!15}{43.4} \\
    \bottomrule
    \end{tabular}
    }
    \end{adjustbox}
    \label{tab:main_exp}
\vspace{-0.5em}
\end{table*}

\paragraph{\textbf{Performance training on MGACorpus}}
We evaluate whether incorporating MGA data enhances model performance compared to a baseline trained solely on the original sources, using fixed training budgets and model sizes ranging from 134M to 1.7B. As shown in~\autoref{tab:main_exp}, MGA-Expansion shows consistent improvements across different model sizes,
with larger performance gains as model size increases, +0.26/+0.95/+2.15 for 134M/377M/1.7B models respectively.
Notably, MGA-Expansion achieved substantial gains in reasoning-intensive tasks such as TriviaQA (+2.03/+6.99/+15.47) and GSM8K (+0.15/+0.22/+6.06), and shows strong performance on MMLU-Pro (all metrics in \autoref{tab:main_exp}). We hypothesize that MGA's data reformulation, by exposing the model to diverse phrasings of the same underlying information, fosters more robust generalization. This enhanced generalization, in turn, enhances the model's reasoning capabilities, leading to the  improvements observed on these specific benchmarks.
For details on the baseline and MGA-Expansion data recipes, including context for comparisons with other models like SmolLM, please refer to \autoref{sec:appd_training}. Additional insights and explanations regarding metric changes are provided in \autoref{sec:appd_experiments}.

\paragraph{\textbf{Scaling Dynamics}}
We further investigate MGA's behavior under data-constrained scaling scenarios. Specifically, we train models ranging from 377M to 13B parameters with only learning rate warmup and stable phase, enabling direct comparison of performance metrics across repetition epochs.

\begin{figure}[h]
  \centering
  \includegraphics[width=\linewidth]{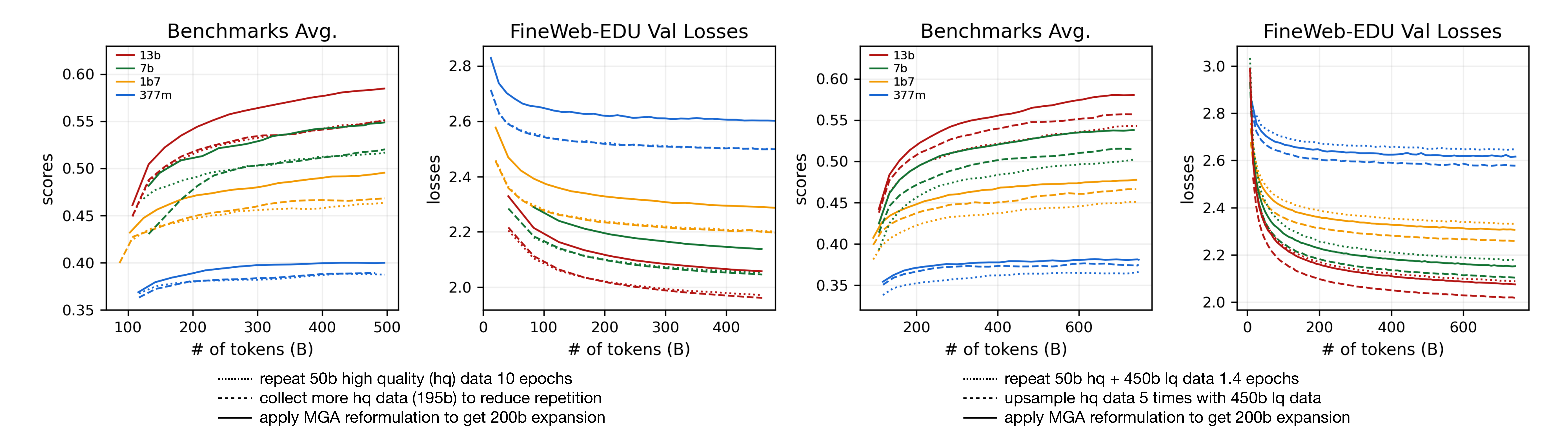}
  \vspace{-2em}
  \caption{Training dynamics of two common scenarios under data-constrained conditions: (1) expanding a 50B high-quality dataset to a 500B training budget (entire set repetition), (2) expanding a 500B mixed-quality dataset to a 700B training budget (subset repetition). 
  For data recipe details please refer to \autoref{sec:appd_training} and benchmark details are provided in Appendix~\ref{sec:appd_scaling_details}.}
  \label{fig:scaling_exp}
  \vspace{-0.5em}
\end{figure}

\vspace{-1em}
\paragraph{\textbf{Scaling Results}}
As shown in~\autoref{fig:scaling_exp}, MGA demonstrates favorable scaling properties with both data budget (D-scaling) and model parameters (N-scaling).
\begin{itemize}[leftmargin=1.5em]
\item In the EntireSet experiments (expanding a 50B high-quality dataset to a 500B budget), simply increasing unique token count by collecting more high-quality data (195B via Full-Fineweb-Edu) shows marginal improvements (+0.2/+0.15/-0.16/+0.11) at 200/300/400/500 billion token steps (13B size). In contrast, MGA, through a 200B reformulation as expansion of the original 50B data, demonstrates consistent gains (+2.65/+3.14/+3.43/+3.46), highlighting \textbf{effective D-scaling}. 
\item Similarly, in the Subset experiments (expanding a 500B mixed-quality dataset to a 700B budget), both upsampling the high-quality portion (5x) and MGA (via a 200B expansion) improve upon the baseline. However, their N-scaling properties with model parameters differ significantly: the performance advantage of upsampling remains relatively constant across model sizes (+0.89/+1.53/+1.23/+1.41), whereas MGA expansion exhibits \textbf{superior N-scaling}, its performance gains amplifying with increasing model scale (+1.46/+2.67/+3.59/+3.73).
\end{itemize}
These scaling experiments demonstrate that our method effectively serves as a data augmentation strategy to mitigate data repetition and aids model scaling (both N and D) in data-constrained scenarios, thus robustly supporting a key aspect of \textbf{RQ1}. While these results highlight MGA's effectiveness, the specific role of its inherent diversity in these high-repetition settings (relevant to \textbf{RQ2}) will be explored further in Section~\ref{sec:discussions}.

\vspace{-0.5em}

\paragraph{\textbf{Validation Losses}}
Although MGACorpus demonstrates superior benchmark performance, we observe increasing validation losses compared to baseline models.
While higher validation losses might seem concerning at first glance, it's important to note that validation loss may not fully reflect model performance,
as token-level perplexity is inherently biased by the frequency distribution of the validation set, and in-domain validation metrics may not necessarily correlate with out-of-domain generalization capabilities.
This observation, combined with recent studies linking loss degradation to model collapse~\citep{dohmatob2024talecollapse,dohmatob2024strongcollapse,zhu2024synthesize}, calls for a more nuanced analysis, which we provide in Section~\ref{sec:discuss_model_collapse}.


\subsection{Discussions}
\label{sec:discussions}
\subsubsection{How effective is reformulation?}
\vspace{-0.5em}
The main experiments in Section~\ref{sec:main_exp} have already established the overall effectiveness of MGA reformulation as a pretraining data augmentation strategy, particularly in data-constrained scenarios. To further solidify this aspect of \textbf{RQ1} and contextualize MGA's performance, we extend our analysis by comparing MGA-enhanced training data with other prominent open-source synthetic datasets: Cosmopedia~\citep{allal2024SmolLM} and variants from the Nemotron family~\citep{su2024nemotron}. This comparison aims to evaluate how MGA's specific reformulation approach—targeting diverse genre and audience presentations of source material—stands against alternative synthetic data generation techniques.

\begin{table*}[hbt!]
    \center
    \vspace{-0.5em}
    \caption{Comparative benchmark performance of 377M models trained on MGA reformulations versus other synthetic datasets for 300B tokens. For a fair comparison with Cosmopedia, MGA is sampled to 28B unique tokens, with both datasets then repeated 10.7 times during training. All benchmarks are 0-shot evaluations (obtained through \textsc{LightEval}), except for MMLU (5-shot). }
    \vspace{-1em}
    \renewcommand{\arraystretch}{1.15}
    \begin{adjustbox}{max width=\textwidth}
    \setlength{\tabcolsep}{0.9mm}{ 
    \begin{tabular}{@{\extracolsep{\fill}}l p{3cm} p{4cm} c c c c c c c c c} 
    \toprule
    Category & Document Sources & Synthetic Target & ARC(C+E) & Wino. & Hella. & MMLU & CSQA & OpenBookQA & PIQA & TriviaQA & Avg. \\
    \midrule
    Cosmopedia & Textbooks/Webs & Story/Textbook/Wiki mix & 42.15 & 50.43 & 45.06 & 29.17 & 30.38 & 33.2 & 68.77 & 0.23 & 35.57 \\
    MGA & High quality webs & Diverse Genre-Audience & 45.65 & 51.22 & 42.31 & 31.42 & 32.19 & 37.2 & 68.39 & 3.79 & 37.28 \\
    \midrule
    \multirow{6}{*}{Nemotron-CC} & Low quality webs & Wrap-medium (Wiki style) & 29.01 & 50.83 & 38.36 & 26.29 & 29.32 & 32 & 67.03 & 0 & 31.72 \\
    &  \multirow{5}{*}{High quality webs} & Extract knowledge & 40.42 & 53.2 & 44.65 & 30.57 & 28.99 & 35 & 69.42 & 0.96 & 35.72 \\
    & & Knowledge list & 42.08 & 52.17 & 42.7 & 30.71 & 32.51 & 35.4 & 70.08 & 0 & 36.21 \\
    & & Concise and clear passage & 42.22 & 52.01 & 43.99 & 30.96 & 31.53 & 35 & 69.7 & 0.06 & 36.21 \\
    & & Wrap-medium (Wiki style) & 42.95 & 52.17 & 43.72 & 31.06 & 31.53 & 36.2 & 70.13 & 0.82 & 36.63 \\
    & & Diverse QA pairs & 46.96 & 52.57 & 49.03 & 31.36 & 38.82 & 38.8 & 70.84 & 9.21 & 40.72\footnotemark \\
    MGA & High quality webs & Diverse Genre-Audience & 45.33 & 52.41 & 42.42 & 31.33 & 31.45 & 38 & 68.61 & 4.24 & 37.34 \\
    \bottomrule
    \end{tabular}
    }
    \end{adjustbox}
    \label{tab:compare_with_synthetic} 
    \vspace{-0.5em}
\end{table*}
\footnotetext{The predominantly 0-shot evaluation particularly benefits datasets like Nemotron `diverse QA pairs' whose format directly aligns with many evaluation tasks.}

The comparative data in \autoref{tab:compare_with_synthetic} highlights MGA's strength. MGA (average 37.28) surpasses Cosmopedia (35.57). Against various Nemotron strategies, MGA (average 37.34) also outperforms most alternatives like `extract knowledge' (35.72) and `wrap-medium (Wiki style)' (36.63). While Nemotron's `diverse QA pairs' achieves the highest average (40.72), its format offers an advantage in the 0-shot evaluation context. Despite this, MGA’s broader reformulation approach demonstrates robust utility, outperforming five of the six Nemotron strategies and showing particular strength on benchmarks like TriviaQA, underscoring its value as a general-purpose data augmentation technique.

These comprehensive comparisons reinforce the answer to \textbf{RQ1}: MGA's diverse genre-audience reformulation is a highly effective pretraining data augmentation strategy. It not only improves upon baselines using original data (as shown in Section~\ref{sec:main_exp}) but also stands as a strong, and often superior, alternative to other synthetic data generation methods. The particular strength of MGA appears to lie in its ability to generate varied and contextually rich reformulations, which likely contributes to the model's enhanced generalization and reasoning capabilities.

\subsubsection{Does reformulation diversity help to mitigate repetition issue?}
\label{sec:ablation_pe} 
\vspace{-0.5em}

To address \textbf{RQ2}, this section examines how different design choices in prompt engineering influence the effectiveness of the MGA framework, particularly under high-repetition conditions.
By comparing SLM variants (introduced in Section~\ref{sec:method_pe}) using different consistency requirements, we identify optimal strategies for balancing information preservation with content diversity.

We sample an additional 20B tokens from real data and generate three synthetic datasets: 80B tokens using SLM-Base, 80B tokens using SLM-Strict, and 40B tokens using SLM-Relaxed. 
Similar to experimental setup in early sections, we set a high-repetition baseline on a smaller data scale (replicating the original 20B tokens 10 times) to more clearly demonstrate the potential impact of SLM-Strict compared to SLM-Base.

\begin{figure}[h]
    \centering
    \vspace{-0.5em}
    \includegraphics[width=\linewidth]{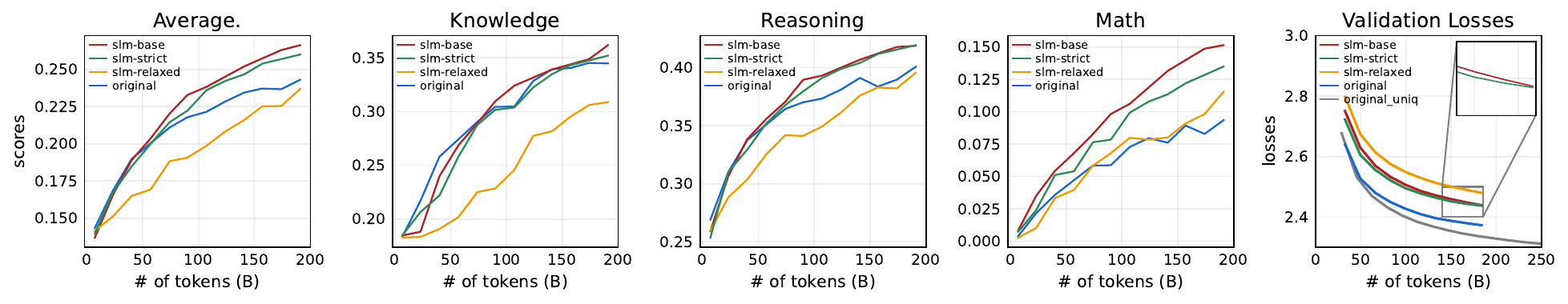}
    \vspace{-2em}
    \caption{Benchmark results and validation losses. The sensitivity to data repetition varies across capability domains, with knowledge dimension showing greater resilience.}
    \label{fig:ablation_pe_training}
\end{figure}
\vspace{-0.5em}

As shown in~\autoref{fig:ablation_pe_training}, our experiments reveal distinct patterns across training configurations.
Both SLM-Base and SLM-Strict demonstrate performance improvements, while the SLM-Relaxed configuration leads to significant collapse.
More details could be found in Appendix~\ref{sec:appd_ablation_details}.

Despite the apparent effectiveness of strict information preservation,
can it fundamentally address the challenges of data repetition?
Our examination of validation loss trajectories reveals a critical distinction:
SLM-Base maintains healthy optimization characteristics throughout training,
whereas SLM-Strict exhibits degraded scaling behavior at higher iteration steps,
reminiscent of the limitations observed with data repetition.
Therefore, this investigation into prompt engineering variants concludes that a balanced `Limited Consistency' approach (SLM-Base) yields best reformulation quality and subsequent model performance answering to \textbf{RQ2}.

\subsubsection{How pretraining data reformulation benefits?}
\label{sec:discuss_model_collapse}
\vspace{-0.5em}

Having explored the impact of reformulation diversity in addressing data repetition (\textbf{RQ2}), we now turn to \textbf{RQ3}: Why does MGA reformulation benefit pretraining performance? We investigate the underlying mechanisms by analyzing learning characteristics and validating against potential issues like model collapse~\citep{dohmatob2024talecollapse,dohmatob2024strongcollapse,zhu2024synthesize}.

\begin{figure*}[h]
\vspace{-0.5em}
\centering
\includegraphics[width=\linewidth]{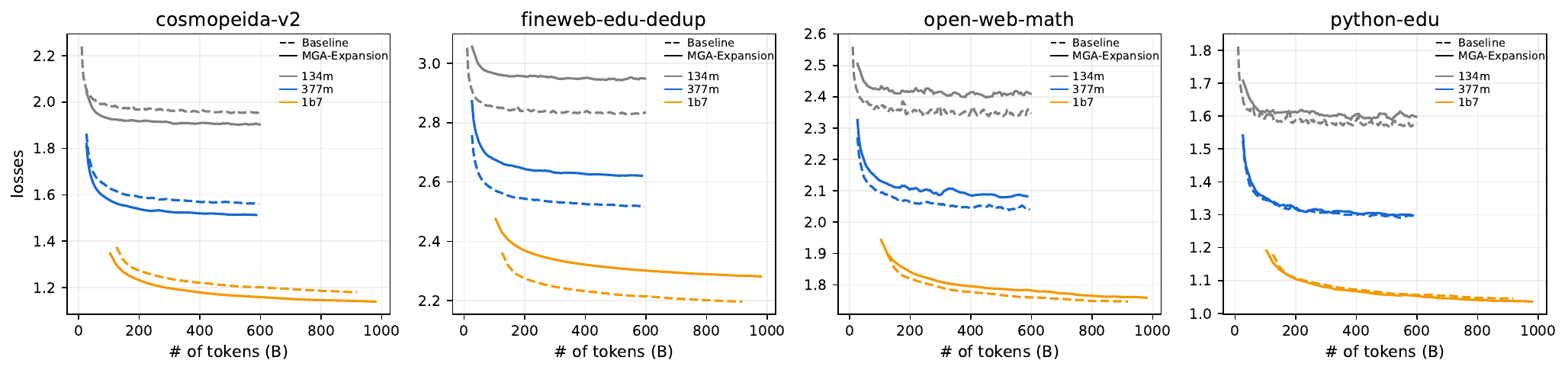}
\vspace{-2em}
\caption{validation losses of experiments in Section~\ref{sec:main_exp}.}
\label{fig:val_loss}
\vspace{-1em}
\end{figure*}

\paragraph{\textbf{Multi-perspective Validation Analysis}}
Our analyses across different validation sets reveal varying patterns in model behavior (\autoref{fig:val_loss}).
As expected, MGA groups' substitution of fineweb-edu data results in adverse effects on corresponding loss, with similar deterioration observed in open-web-math.
Interestingly, the synthetic dataset cosmopedia demonstrates improved loss metrics.
A notable contrast emerges in python-edu: while MGA exhibit negative impact at the 134M and 377M parameter, this trend reverses at 1.7B, suggesting scale-dependent effects on model behavior.

\begin{figure*}[h]

    \centering
    \includegraphics[width=\linewidth]{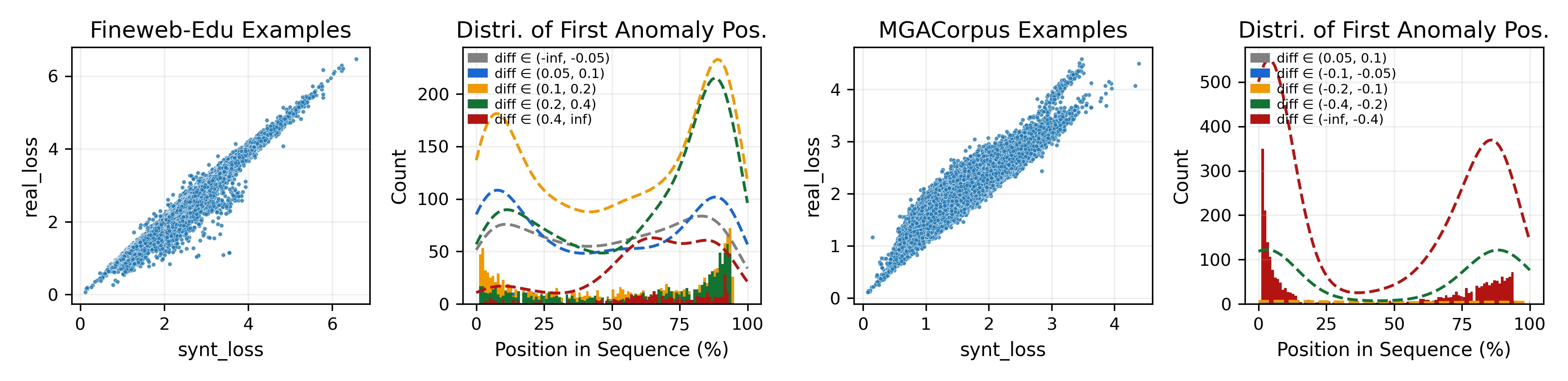}
    \vspace{-2em}
\caption{Losses pattern analysis. Subfigures 1 and 3 shows comparison between models trained on different data settings, with $loss_{\text{real}}$ on y-axis and $loss_{\text{synt}}$ on x-axis.
Subfigures 2 and 4 track the position where $loss^i_{\text{synt}}-loss^i_{\text{real}}$ ($loss^i_{\text{diff}}$) first becomes significantly higher than the sequence's average difference (detailed definition in Appendix~\ref{sec:appd_ablation_details}).}
\label{fig:doc_loss}
\end{figure*}

\paragraph{\textbf{Fine-grained Pattern Analysis}}
To better understand whether increased validation loss truly indicates model collapse,
we conduct a fine-grained analysis of loss patterns.
Specifically, we compare token-level losses of 800B checkpoint between models trained on real data and synthetic data (Baseline and MGA-Expansion in Section~\ref{sec:main_exp}, respectively). The document samples are from both Fineweb-Edu and MGACorpus.
As illustrated in subfigures 1 and 3 of \autoref{fig:doc_loss},
each point represents a sample's average token loss,
consistent with the overall loss discrepancy shown in \autoref{fig:val_loss}.

The distribution of first anomaly positions (subfigures 2 and 4) reveals a crucial insight:
when processing real data, models trained on synthetic data show performance degradation (measured by $loss_{\text{diff}}$) that predominantly manifests in later sequence positions,
which intensifies as $loss_{\text{diff}}$ increases.
However, this positional bias disappears when evaluating on synthetic data.

The systematic pattern suggests that rather than experiencing model collapse,
the synthetic-trained model may have developed a different learning strategy (cases in Appendix~\ref{sec:appd_ablation_details}).
While it shows higher validation losses on certain real-world datasets,
its strong performance in our main experiments indicates a potential trade-off:
\textbf{the model may prioritize learning generalizable patterns from context over memorizing specific sequence dependencies.}
This shift in learning process could explain both the improved performance on benchmarks
and increased losses on validation sets that potentially require more memorization-based processing.

Addressing \textbf{RQ3}, these findings indicate that the performance characteristics associated with MGA data likely stem from altered learning strategies, potentially prioritizing generalizability, rather than representing model collapse phenomenon.

\section{Conclusion}
\vspace{-0.5em} 
In this work, we introduced MGA, an efficient framework that leverages genre-audience reformulation to systematically expand existing corpora with diverse, synthetically generated variations. 
Our core finding highlights MGA's effectiveness as a data augmentation strategy specifically targeting the repetition challenge: in data-constrained scaling experiments, MGA significantly outperformed naive data repetition and simple upsampling, enabling more effective model training beyond unique data limits. 
Furthermore, the quality of the MGACorpus was confirmed by consistent performance improvements when incorporated into standard training mixtures across various model sizes. 
While evaluating synthetically expanded data requires careful consideration, MGA's success stems from its ability to create relevant diversity, directly counteracting the negative impacts of repeating limited datasets. 
Therefore, MGA offers a practical and scalable pathway to alleviate data repetition bottlenecks, facilitating continued progress in large language model development by making more effective use of available data resources.

\subsubsection*{Acknowledgements}
We are grateful to Chao He, Zhixin Yao, Yue Chen and Runyu Shi for their help with prompt templates and case studies,
and to Seed-Foundation team for providing the stable training/inference platform, 
which enabled us to build the synthetic pipeline and corpus within a reasonable timeframe.
The icons shown in Figure 1 are designed by Freepik.


\bibliographystyle{unsrtnat}
\bibliography{main}

\clearpage
\beginappendix
\section{Limitations and Opportunities}
\label{sec:appd_limitations}
\vspace{-0.5em}

While our experimental results demonstrate the effectiveness of MGA in both quality validation and scaling scenarios, several important aspects warrant further investigation. We identify three key areas:
\begin{itemize}[leftmargin=1.5em]
    \item The tool model (SLM) employed in this work relies on an early version with relatively moderate capabilities. While~\citet{pieler2024rephrasing} suggests that model size may not be the determining factor for rephrasing tasks, the influence within MGA framework remains unexplored. Understanding the relationship between SLM capacity and corpus quality is crucial for optimizing the effectiveness-efficiency trade-off.
    
    \item Our current experiments demonstrate effectiveness up to 13B parameters and 1,000B tokens of training budget. Extending this approach to long-horizon training and larger-scale models requires additional validations, particularly for next-generation models which require hundreds of trillions of training tokens.
    
    \item Regarding data repetition strategies, we present preliminary explorations under computational resource constraints. The underlying patterns and their sensitivity to various factors, such as repetition ratio, data distribution, and data quality, require systematic investigation. Future research should examine how these factors collectively determine optimal data strategies across different training scenarios.
\end{itemize}

\paragraph{\textbf{Broader Impact}}
This paper explores the use of LLMs as a data expansion method for pretraining large language models. We introduce a lightweight and scalable framework (MGA) to mitigate data repetition issues, which holds potential for positive societal impact, particularly in synthetic data generation for training language models. Nonetheless, the use of synthetic data generated by LLMs is not without risks; for instance, LLM hallucinations, even after filtering, could introduce novel errors or biases into models trained on such data, a factor that warrants careful consideration in future research and deployment.

\section{Tool Model Implementation}
\label{sec:appd_implementation_details}

\begin{figure}[h]
    \centering
    \vspace{-1.5em}
    \includegraphics[width=\columnwidth]{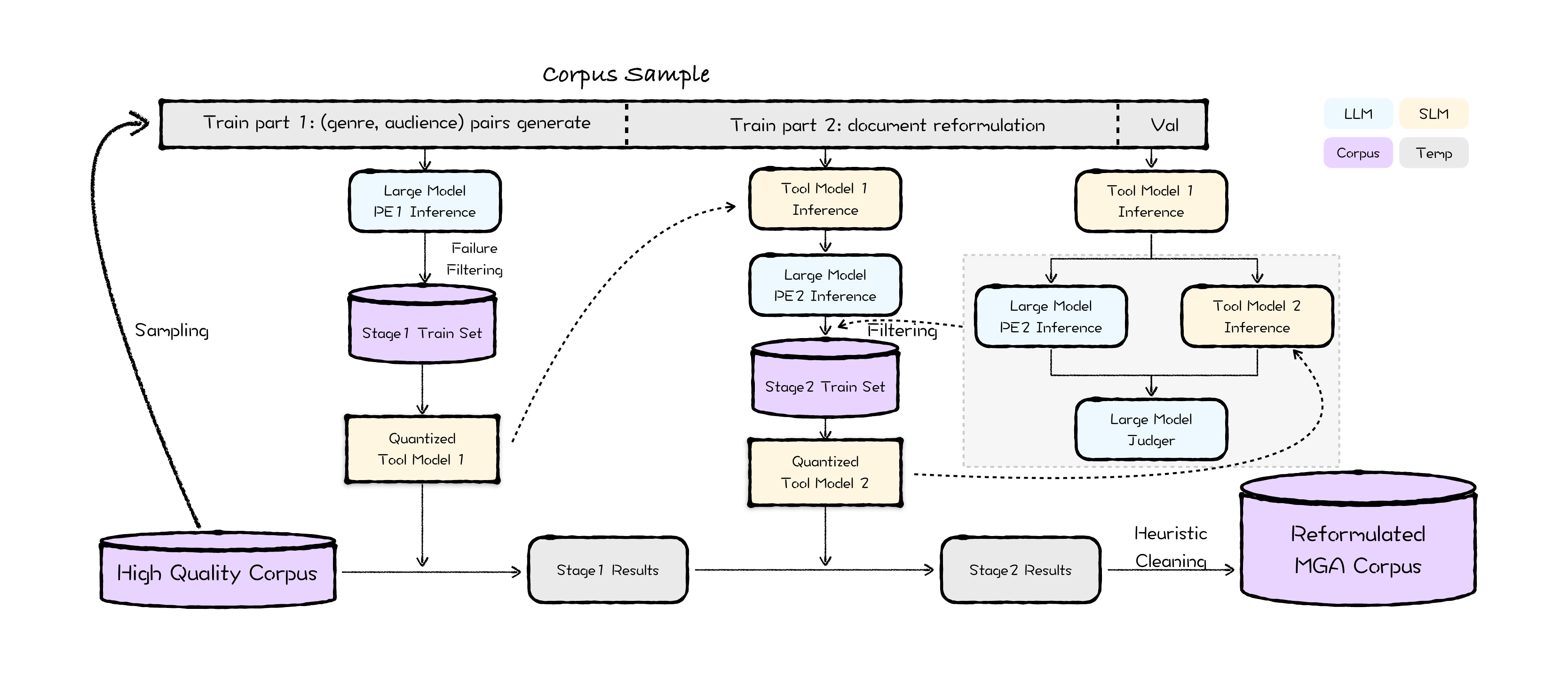}
    \vspace{-2em}
    \caption[width=0.8\columnwidth]{Implementation details. From a high-quality corpus, we sample a subset to serve as input for the LLM labeler and judger. 
    Through iterative filtering, we train and quantize SLM tool models for each stage to improve inference efficiency,
    which are used to generate the reformulated corpus.}
    \label{fig:synthesis-pipeline}
    \vspace{-1em}
\end{figure}

\paragraph{\textbf{High Quality Corpus}}
To ensure reproducibility, we conduct our reformulated corpus based on SmolLM-Corpus\footnote{\url{https://github.com/huggingface/smollm/tree/main/text/pretraining}}~\citep{benallal2024smollmcorpus}, expanding fineweb-edu-dedup source from 195B tokens to 770B tokens.
Then we setup additionally experiments on FineWeb and FineWeb-Edu~\citep{penedo2024finewebdatasets},
which constitute a solid foundation for research on data scaling approaches.
Prior to these experiments, we have validated the approach on our in-house datasets.
The results demonstrate strong performance across both datasets, suggesting broad applicability of our method.

\paragraph{\textbf{Tool Models Training}}
Initialized from a pretrained SLM (a 3.3B MoE model), we collect 50,000 training samples through iterative filtering and training, where 15,000 of raw text to genre-audience pairs, 35,000 of raw text to reformulated output.
Each model's validation responses are scored by capable LLM judger, that ensures the SLMs achieve comparable synthesis quality to the LLM labeler as shown in~\autoref{table:tool_model_scores}.
The sequence length is 8192 with maximum prompt/response length 4,096 tokens, each model is trained 3 epochs on the samples with a cosine lr scheduler.

\paragraph{\textbf{Cleaning Stage}}
Similar to previous synthesis work~\citep{su2024nemotron,maini2024rephrasing}, we involve a final cleaning stage to filter out the high frequency patterns, for example,
`\textit{Notes: ...}', `\textit{Please note that ...}', `\textit{The above is as required ...}', `\textit{The following is...}', etc.
And remove documents with an extremely low keyword coverage to raw documents.

\paragraph{\textbf{Resource Analysis}}
To generate 770B synthetic tokens, it takes 256×64 and 1024×130 NVIDIA H100 GPU hours to process two stages, or 4× more hours when using Huawei Ascend910B2. 
In our practice, we use Huawei Ascend910B2 synthesis most tokens of MGACorpus, which significantly reduce the cost of synthesis.

\section{Pretraining Details}
\label{sec:appd_training}
\vspace{-0.5em}

\paragraph{\textbf{Data Recipe}}
The training token budgets are 600B/600B/1000B for size of 134M/377M/1.7B models, which are aligned with SmolLM1 series~\citep{benallal2024smollmcorpus}.
Our baseline is trained on SmolLM-Corpus dataset, in contrast to SmolLM's recipe, we use unique token number from each source as the mixing ratio shown in \autoref{table:subsource_weight}.
This ensures that different sources have consistent repetition epochs during training. For a fair comparison, the mixing ratios of other data sources are kept constant across experiments. We specifically adjusted the proportions of fineweb-edu-dedup and MGACorpus (which is derived from fineweb-edu-dedup) to isolate the impact of the MGA reformulation.

\begin{table*}[h]
  \vspace{-0.5em}
  \centering
  \caption{MGACorpus experiments data recipe: source weight (\%) and \#unique\_tokens × \#epochs (using 1000B budget as example).
  }
  \vspace{-0.5em}
  \begin{adjustbox}{max width=0.95\textwidth}
    \setlength{\tabcolsep}{1mm}{
  \begin{tabular}{lcccccc}
  \toprule
  experiments & - & fineweb-edu-dedup & cosmopedia-v2 & python-edu & open-web-math & MGACorpus \\
  \midrule
  Baseline & weight & 80.89 & 11.65 & 1.66 & 5.80 & - \\
  & \#unique\_tokens × \#epochs & 195 × 4.15 & 28 × 4.15 & 4 × 4.15 & 14 × 4.15 & - \\
  \midrule
  MGA-Expansion & weight & \textbf{16.29} & 11.65 & 1.66 & 5.80 & \textbf{64.59} \\
   & \#unique\_tokens × \#epochs & 195 × 0.84 & 28 × 4.15 & 4 × 4.15 & 14 × 4.15 & 770 × 0.84 \\
  \bottomrule
  \end{tabular}
  }
  \end{adjustbox}

  \label{table:subsource_weight}
  \vspace{-0.5em}
\end{table*}

The experiment design for different strategies is presented in \autoref{table:scaling_weight},
which involves three datasets:
(1) a 50B-token random sample from fineweb-edu-dedup,
(2) a corresponding filtered subset from MGACorpus,
and (3) a 450B-token deduplicated corpus obtained from Fineweb~\citep{penedo2024finewebdatasets}.

\begin{table*}[h]
  \vspace{-0.5em}
  \centering
  \caption{Scaling experiments data recipe, values represent \#unique\_tokens × \#epochs.}
  \vspace{-0.5em}
  \begin{adjustbox}{max width=\textwidth}
    \setlength{\tabcolsep}{1mm}{
  \begin{tabular}{ccccccl}
  \toprule
  \multirow{2}*{Repetition} &\multirow{2}*{Experiments} & Training  & fineweb-edu & MGA & fineweb & \multicolumn{1}{c}{\multirow{2}*{Design Rationale}} \\
  & & Budget & dedup & corpus & random & \\
  \midrule
  \multirow{3}*{EntireSet}  & Baseline         & 500B & 50 × 10 & - & -  & \\
                            & Full-Fineweb-Edu & 500B & 195 × 2.56 & - & - & What if we could collect more unique data.\\
                            & MGA Expansion   & 500B & 50 × 2 & 200 × 2 & - & Add MGA to reduce the repetition num. \\
  \midrule
  \multirow{3}*{Subset } & Baseline       & 700B & 50 × 1.4 & - & 450 × 1.4 & \\
                         & Upsample-EDU   & 700B & 50 × 5 & - & 450 × 1 & Upsample to get 200B more budget.\\
                         & MGA Expansion & 700B & 50 × 1 & 200 × 1 & 450 × 1 & Add MGA to achieve the same target.  \\
  \bottomrule
  \end{tabular}
  }
  \end{adjustbox}

  \label{table:scaling_weight}
\end{table*}

\paragraph{\textbf{Training Hyperparameters}}
We sample 100 million tokens from SmolLM-Corpus as the validation dataset. 
The hyperparams are presented in \autoref{tab:model_hyperparams}, tokenzier used for training and computing token counts is same as 
SmolLM1\footnote{\url{https://huggingface.co/HuggingFaceTB/cosmo2-tokenizer}} with a vocab size of 49,152.

\paragraph{\textbf{Evaluation}}
The LightEval results provided in Section~\ref{sec:main_exp} follow SmolLM setting, that with GSM8K/MMLU 5-shot and all the others 0-shot.
The benchmarks presented in \autoref{fig:appd_benchs1} and \autoref{fig:appd_benchs2} follow few-shot evaluation settings, specifically ARC(8-shots), TriviaQA(5-shots), Winogrande(5-shots) 
and similar configurations for other tasks.

\begin{table*}[t]
    \centering
    \setlength{\tabcolsep}{4pt}
    \vspace{-2em}
    \caption{Hyperparams of different model size.}
    \vspace{-0.5em}
    \begin{adjustbox}{max width=0.9\textwidth}
    \begin{tabular}{*{11}{c}}
        \toprule
        model & batch & learning & hidden & ffn & num & num & shared & seq & tie & total \\
        size & size & rate & size & inner & heads & layers & q\_head & len & emb & params \\
        \midrule
        \textbf{134M} & 128  & 3e-3   & 1,204 &  4,096 &  8 &  8 & 1 & 8,192 & false & 134M \\
        \textbf{377M} & 320  & 1.5e-3 & 1,536 &  6,144 & 12 & 10 & 1 & 8,192 & false & 377M \\
        \textbf{1.7B} & 512  & 5e-4   & 2,560 & 10,240 & 20 & 16 & 1 & 8,192 & false & 1.68B \\
        \textbf{7B}  & 1,024 & 4e-4   & 4,096 &  8,192 & 32 & 32 & 4 & 8,192 & false & 6.98B \\
        \textbf{13B} & 1,024 & 4e-4   & 4,096 & 12,288 & 32 & 48 & 4 & 8,192 & false & 12.9B \\
        \bottomrule
    
    \end{tabular}
    \end{adjustbox}
    \label{tab:model_hyperparams}
    \vspace{-1em}
\end{table*}

\section{Further Analysis of Experiments}
\label{sec:appd_experiments}
\subsection{Benchmark Improvement}
\vspace{-0.5em}

In our experimental observations (\autoref{tab:appd_bench_analysis}), notable performance improvements are demonstrated in both TriviaQA and GSM8k benchmarks,
 warranting a detailed examination of these score variations. 

\begin{table*}[h]
    \center
    \setlength{\fboxsep}{1pt}
    \caption{Benchmark results. A copy of SmolLM1/SmolLM1-Ours/MGA-Expansion in \autoref{tab:main_exp}.
    }
    \vspace{-0.5em}
    \renewcommand{\arraystretch}{1.15}
      
    \begin{adjustbox}{max width=\textwidth}
    \setlength{\tabcolsep}{1mm}{
    \setlength{\fboxsep}{1.5pt}
    \begin{tabular}{@{\extracolsep{\fill}}lcccccccccccccccc}
    \toprule
      
      Model & \#Params. & \#Tokens & ARC(C+E) & Wino. & Hella. & MMLU & MMLU-PRO & CSQA & OpenBookQA & PIQA & TriviaQA & GSM8K & Avg. \\
      \midrule
      SmolLM-1.7B & 1.7B & 1T & 59.95 & 54.7 & 62.83 & 39.35 & 10.92 & 38 & 42.6 & 75.9 & 13.14 & 4.62 & 40.20 \\
      Baseline & 1.7B & 1T & 59.63 & 57.38 & 65.19 & 39.4 & 12.11 & \underline{42.59} & \textbf{45.6} & 76.88 & 4.95 & 7.81 & 41.15 \\
      MGA-Expansion & 1.7B & 1T & \colorbox{green!15}{\underline{60.36}} & \colorbox{green!15}{57.46} & \colorbox{green!15}{65.52} & \colorbox{green!15}{\underline{40.79}} & \colorbox{green!15}{\underline{14.1}} & 41.11 & \underline{42.8} & \colorbox{green!15}{77.53} & \colorbox{green!15}{20.42} & \colorbox{green!15}{13.87} & \colorbox{green!15}{43.4} \\
    \bottomrule
    \end{tabular}
    }
    \end{adjustbox}
    \label{tab:appd_bench_analysis}

\end{table*}
The enhanced TriviaQA performance exhibited by SmolLM1-1.7B relative to our baseline 
can be attributed to the larger proportion of Cosmopedia in its training configuration.
Both MGACorpus and Cosmopedia employ synthetic methodologies, which contribute to improved learning efficiency. 
The observed gains in GSM8K performance can be traced to the target genres, including teaching schemas and problem-solving exemplars, embedded within the Reformulation component. 
This early exposure to structured problem-solving approaches facilitates more effective performance on analogous mathematical reasoning tasks.

\subsection{What if use MGACorpus alone?}
Our primary goal with MGA is efficient dataset expansion, typically achieved by mixing the generated corpus with existing real data, aligning with current best practices for leveraging synthetic data. However, to better characterize the properties of the MGACorpus itself and understand the impact of training exclusively on reformulated content, we also investigate an experimental setting where MGACorpus completely replaces its source data (fineweb-edu-dedup).
\label{sec:appd_magaonly}
\begin{table*}[h]
  \centering
  \caption{MGACorpus experiments data source weight (\%).
  }
  \begin{adjustbox}{max width=0.85\textwidth}
    \setlength{\tabcolsep}{1mm}{
  \begin{tabular}{lccccc}
  \toprule
  experiments & fineweb-edu-dedup & cosmopedia-v2 & python-edu & open-web-math & MGA-corpus \\
  \midrule
  Baseline & 80.89 & 11.65 & 1.66 & 5.80 & - \\
  MGA-Only & - & 11.65 & 1.66 & 5.80 & \textbf{80.89} \\
  MGA-Expansion & \textbf{16.29} & 11.65 & 1.66 & 5.80 & \textbf{64.59} \\
  \bottomrule
  \end{tabular}
  }
  \end{adjustbox}

  \label{table:appd_subsource_weight}
\end{table*}

As shown in~\autoref{tab:main_exp2}, the absence of real data leads to performance degradation across most tasks (average -0.95), 
particularly in two tasks, Hellaswag(-1.23/-1.69/-2.85) and CommonsenseQA(-3.11/-4.83/-4.50). 
This decline can be attributed to our design choice, 
which focuses on diversity and overall quality rather than requiring the preservation of all information from each raw documents.

\begin{table*}[tb!]
  \center
 
  \caption{Comparison between MGA-Expansion and MGA-Only}
  \vspace{-0.5em}
  \begin{adjustbox}{max width=\textwidth}
  \setlength{\tabcolsep}{1mm}{
  \setlength{\fboxsep}{0pt}
  \begin{tabular}{@{\extracolsep{\fill}}lcccccccccccccccc}
  \toprule
    
    Model & \#Params. & \#Tokens & ARC(C+E) & Wino. & Hella. & MMLU & MMLU-PRO & CSQA & OpenBookQA & PIQA & TriviaQA & GSM8K & Avg. \\
    \midrule
    MGA-Expansion & 134M & 600B & \textbf{43.01} & \textbf{51.7} & \textbf{41.25} & \textbf{30.1} & \textbf{11.76} & \textbf{32.68} & \textbf{36.4} & 67.3 & 2.05 & \textbf{1.44} & \textbf{31.77}\\
    MGA-Only & 134M & 600B & 41.98 & 51.38 & 40.02 & 29.87 & 11.5 & 29.57 & 33 & \textbf{68.01} & \textbf{2.26} & 1.06 & 30.87\\
    &  &  & \colorbox{red!15}{$\downarrow$-1.03} & \colorbox{red!15}{$\downarrow$-0.32} & \colorbox{red!15}{$\downarrow$-1.23} & \colorbox{red!15}{$\downarrow$-0.23} & \colorbox{red!15}{$\downarrow$-0.26} & \colorbox{red!15}{$\downarrow$-3.11} & \colorbox{red!15}{$\downarrow$-3.40} & \colorbox{green!15}{$\uparrow$0.71} & \colorbox{green!15}{$\uparrow$0.21} & \colorbox{red!15}{$\downarrow$-0.38} & \colorbox{red!15}{$\downarrow$-0.90}\\
    \midrule
    MGA-Mix & 377M & 600B & \textbf{49.39} & 52.64 & \textbf{51.34} & \textbf{34.09} & 11.35 & \textbf{37.1} & \textbf{38} & \textbf{72.31} & \textbf{7.28} & \textbf{1.74} & \textbf{35.52}\\
    MGA-Only & 377M & 600B & 47.95 & \textbf{53.35} & 49.65 & 33.31 & \textbf{11.38} & 32.27 & 38 & 70.95 & 6.83 & 1.59 & 34.53\\
    &  &  & \colorbox{red!15}{$\downarrow$-1.44} & \colorbox{green!15}{$\uparrow$0.71} & \colorbox{red!15}{$\downarrow$-1.69} & \colorbox{red!15}{$\downarrow$-0.78} & \colorbox{green!15}{$\uparrow$0.03} & \colorbox{red!15}{$\downarrow$-4.83} &  -  & \colorbox{red!15}{$\downarrow$-1.36} & \colorbox{red!15}{$\downarrow$-0.45} & \colorbox{red!15}{$\downarrow$-0.15} & \colorbox{red!15}{$\downarrow$-0.99}\\
    \midrule
    MGA-Mix & 1.7B & 1T & \textbf{60.36} & \textbf{57.46} & \textbf{65.52} & \textbf{40.79} & \textbf{14.1} & \textbf{41.11} & 42.8 & \textbf{77.53} & \textbf{20.42} & \textbf{13.87} & \textbf{43.40}\\
    MGA-Only & 1.7B & 1T & 59.02 & 57.06 & 62.67 & 40.34 & 13.51 & 36.61 & \textbf{45.2} & 76.71 & 19.78 & 13.57 & 42.45\\
    &  &  & \colorbox{red!15}{$\downarrow$-1.34} & \colorbox{red!15}{$\downarrow$-0.40} & \colorbox{red!15}{$\downarrow$-2.85} & \colorbox{red!15}{$\downarrow$-0.45} & \colorbox{red!15}{$\downarrow$-0.59} & \colorbox{red!15}{$\downarrow$-4.50} & \colorbox{green!15}{$\uparrow$2.40} & \colorbox{red!15}{$\downarrow$-0.82} & \colorbox{red!15}{$\downarrow$-0.64} & \colorbox{red!15}{$\downarrow$-0.30} & \colorbox{red!15}{$\downarrow$-0.95}\\
    \bottomrule
  \end{tabular}
  }
  \end{adjustbox}
  \label{tab:main_exp2}
\end{table*}

\subsection{Ablation Details}
\label{sec:appd_ablation_details}
\vspace{-0.5em}
\paragraph{\textbf{MGA-Only Setting of PE Ablation}}
Upon relaxing the information preservation requirements for PE objectives in the MGA-Only setting, 
we observe a complete collapse in knowledge-based dimensions while 
maintaining modest improvements in reasoning and mathematical capabilities. 
This divergence suggests that different cognitive capabilities have distinct requirements 
for the richness and nature of training data content.
\begin{figure*}[htb!]
    \centering
    \vspace{-0.5em}
    \includegraphics[width=\linewidth]{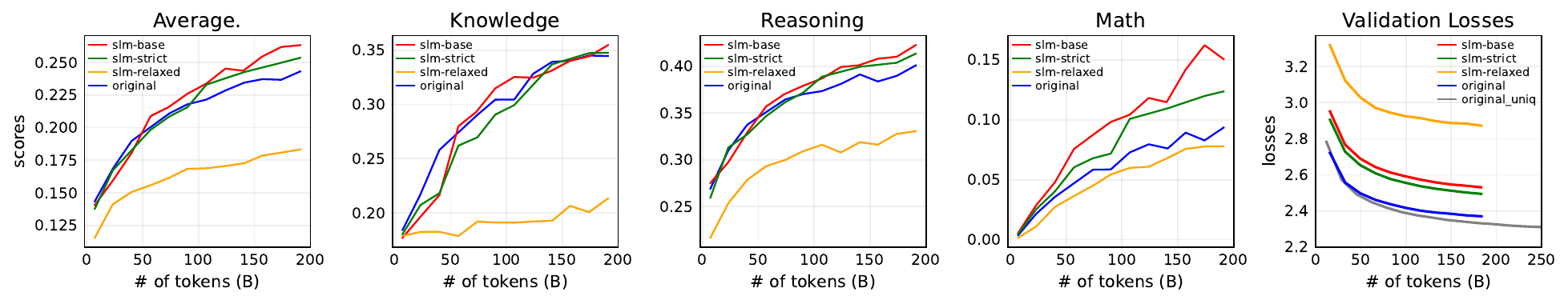}
    \vspace{-2em}
    \caption{Corresponding benchmark results described in Section~\ref{sec:ablation_pe}.}
    \vspace{-1em}
\end{figure*}

\paragraph{\textbf{Further Discussion of Section~\ref{sec:discuss_model_collapse}}}
For our analysis method in \autoref{fig:doc_loss}, we define the token loss difference as $loss^i_{\text{diff}}=loss^i_{\text{synt}}-loss^i_{\text{real}}$, where i is the token index, synt/real is dataset used for model training.
Note that we consistently use synthetic minus real, where a positive value indicates poorer prediction performance by the synthetic model on a given sample.

Since next token prediction is computed based on preceding context, we define the first anomaly position to identify where a model's prediction for tokens within the window begins to significantly deteriorate. 
The definition is as follows:
\begin{align*}
\text{first\_anomaly\_position} = \min\{p ~|~ \left| \frac{1}{w}\sum_{i=p}^{p+w-1} loss^i_{\text{diff}}\right| > |\mu| + k\sigma\},
\end{align*}

where $w = \max(0.05 \times \text{seq\_length}, 1),~\mu = \text{mean}(loss^i_{\text{diff}}),~\sigma = std({loss^i_{\text{diff}}})$.
Here, we employ the absolute value of the windowed average loss to identify significant performance degradation in either model. 
This approach enables the detection of notable prediction quality drops regardless of which model (synthetic or real) experiences the deterioration.

Finally, we define the normalized position, enabling fair comparisons across various sequence lengths:
\begin{align*}
    \text{normalized\_position} = 
    \begin{cases} 
        \frac{\text{first\_anomaly\_position}}{\text{seq\_length}} \times 100\% & \text{if anomaly found} \\ 
        -1 & \text{otherwise} 
    \end{cases}
\end{align*}

\begin{figure*}[htb!]
    \centering
    \includegraphics[width=\linewidth]{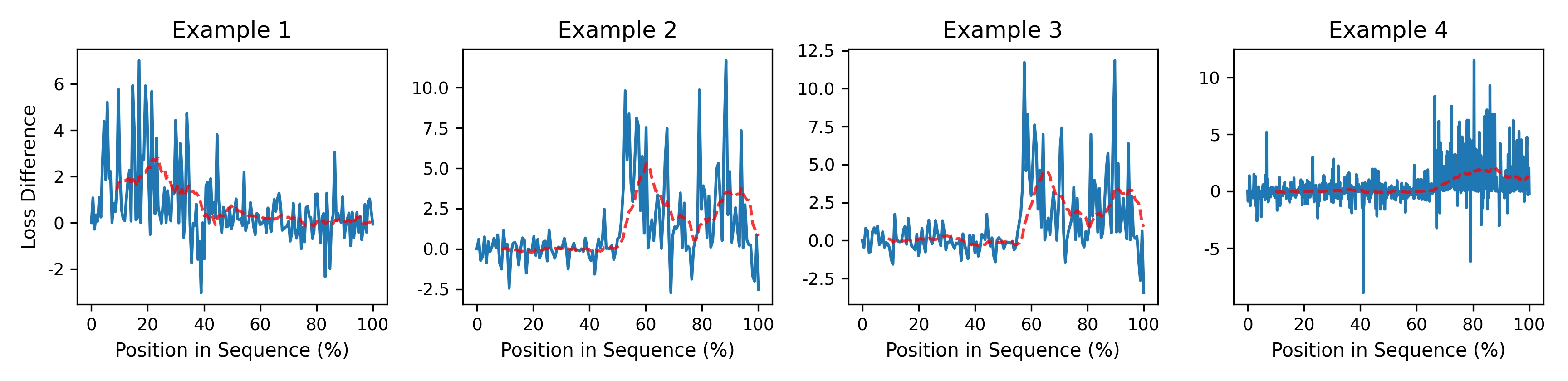}
    \vspace{-1em}
    \caption{Random examples sampling from where $\text{mean}(loss^i_{\text{diff}})>0.5$, the synthetic-trained model fail to predict the tokens in later sequence positions.}
    \label{fig:token_diff_examples}
\end{figure*}

\begin{figure*}[htb!]
    \centering
    \includegraphics[width=\linewidth]{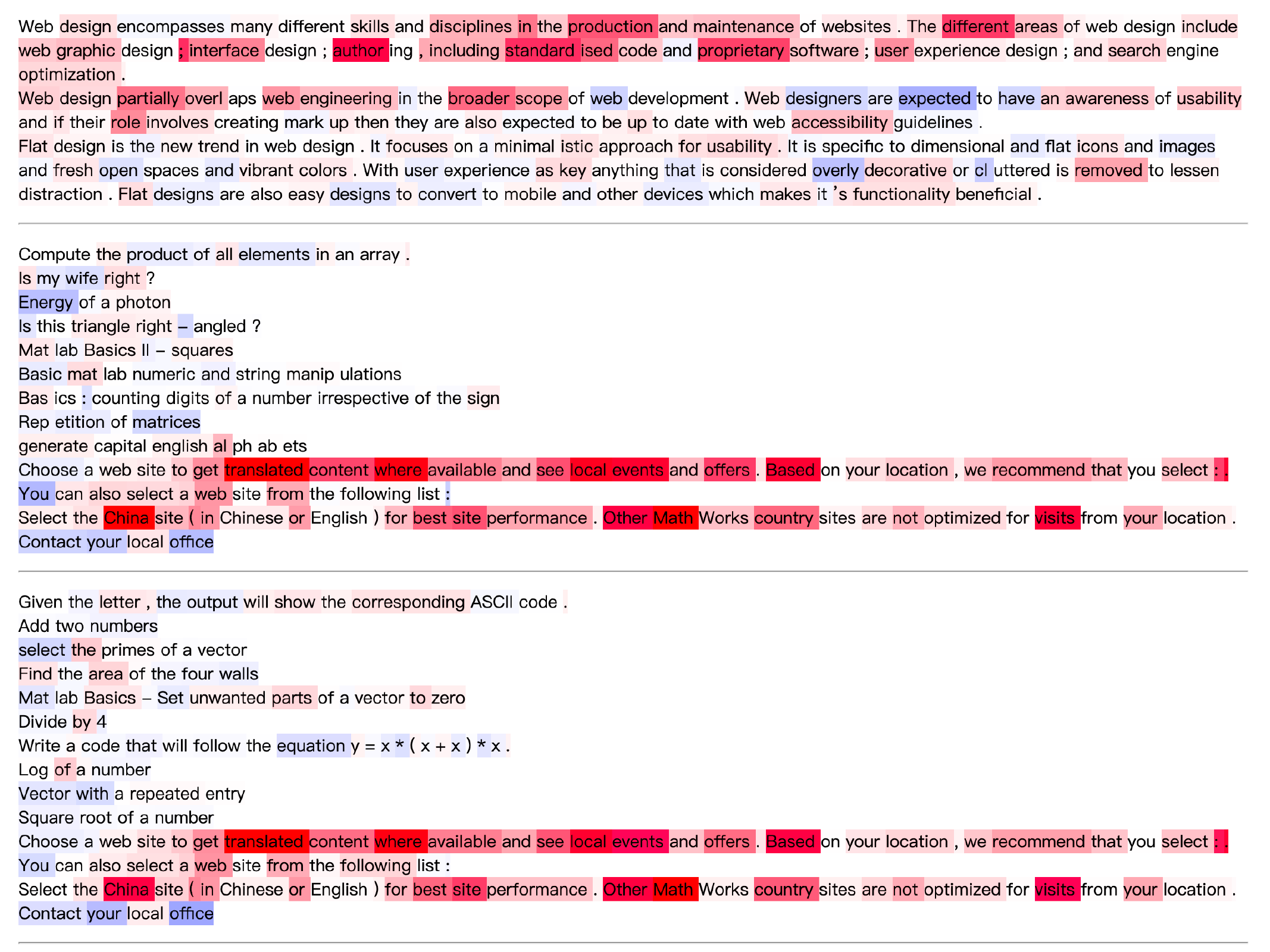}
    \vspace{-2em}
    \caption{Corresponding cases sampled from Fineweb-Edu, which align with the loss patterns shown in Figure~\ref{fig:token_diff_examples}, with higher loss by synthetic-trained model highlighted in \colorbox{red!15}{red}.}
    \label{fig:cases_fineweb}
\end{figure*}

Below are example cases from English and Chinese documents.
\autoref{fig:token_diff_examples} presents the token loss difference on each position.
Example 2 and Example 3 show similar anomaly pattern, we can get the reason in \autoref{fig:cases_fineweb},
that they are from the same website source contain identical boilerplate text about region selection and website localization at the end of their content.

This suggests potential noise in the data preprocessing pipeline, 
specifically in handling website navigation elements and localization prompts 
that should have been removed during content extraction.

While these examples demonstrate clear patterns of model behavior differences in handling noisy web data, we acknowledge that this analysis is limited to selected cases with apparent preprocessing artifacts. A more comprehensive evaluation across diverse data sources and quality levels would be necessary to fully understand the impact of synthetic training data on model performance.

\begin{figure*}[htb!]
    \centering
    \includegraphics[width=\linewidth]{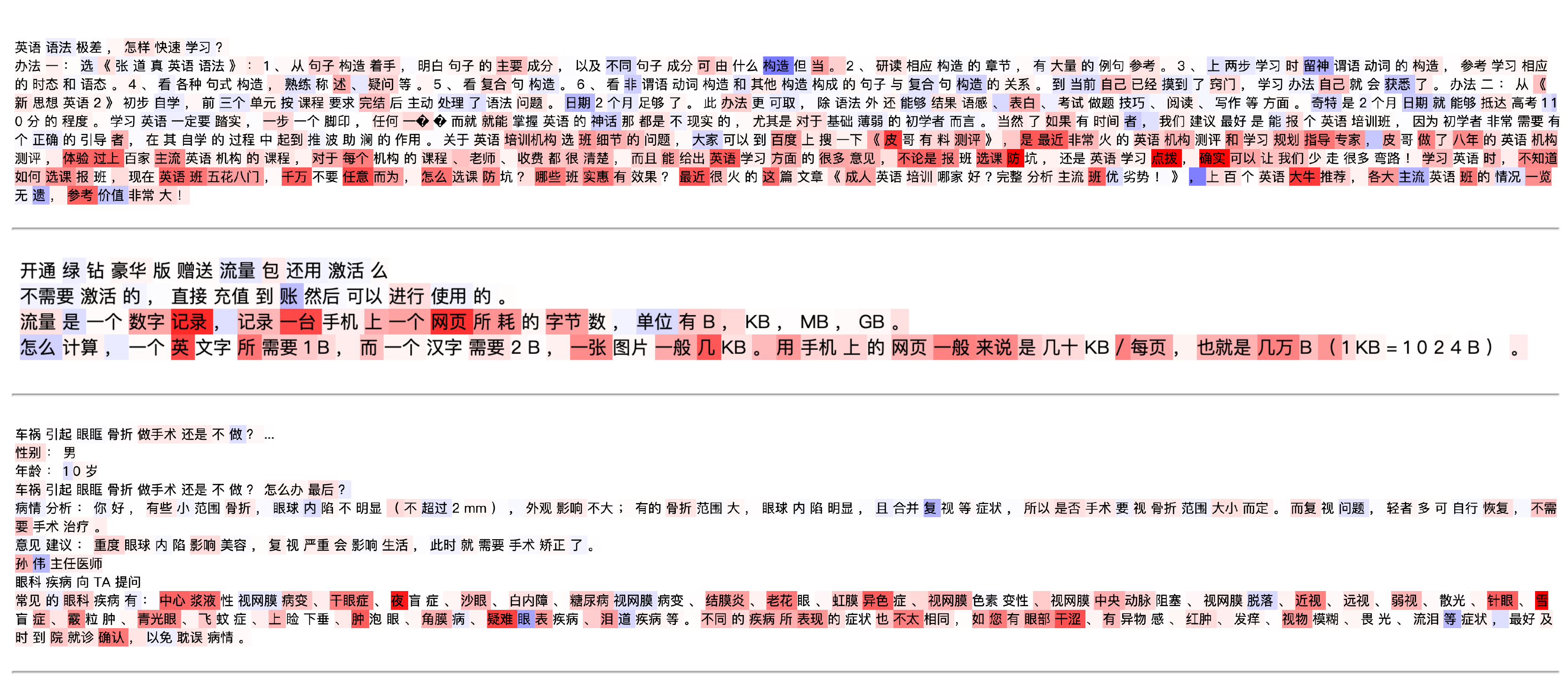}
    \vspace{-2em}
    \caption{Chinese corpus samples with higher loss by synthetic-trained model in \colorbox{red!15}{red}.}
    \label{fig:cases_chinese}
\end{figure*}

\subsection{Scaling Experiments Details}
\label{sec:appd_scaling_details}

\begin{figure*}[htb!]
    \centering    
    \includegraphics[width=\linewidth]{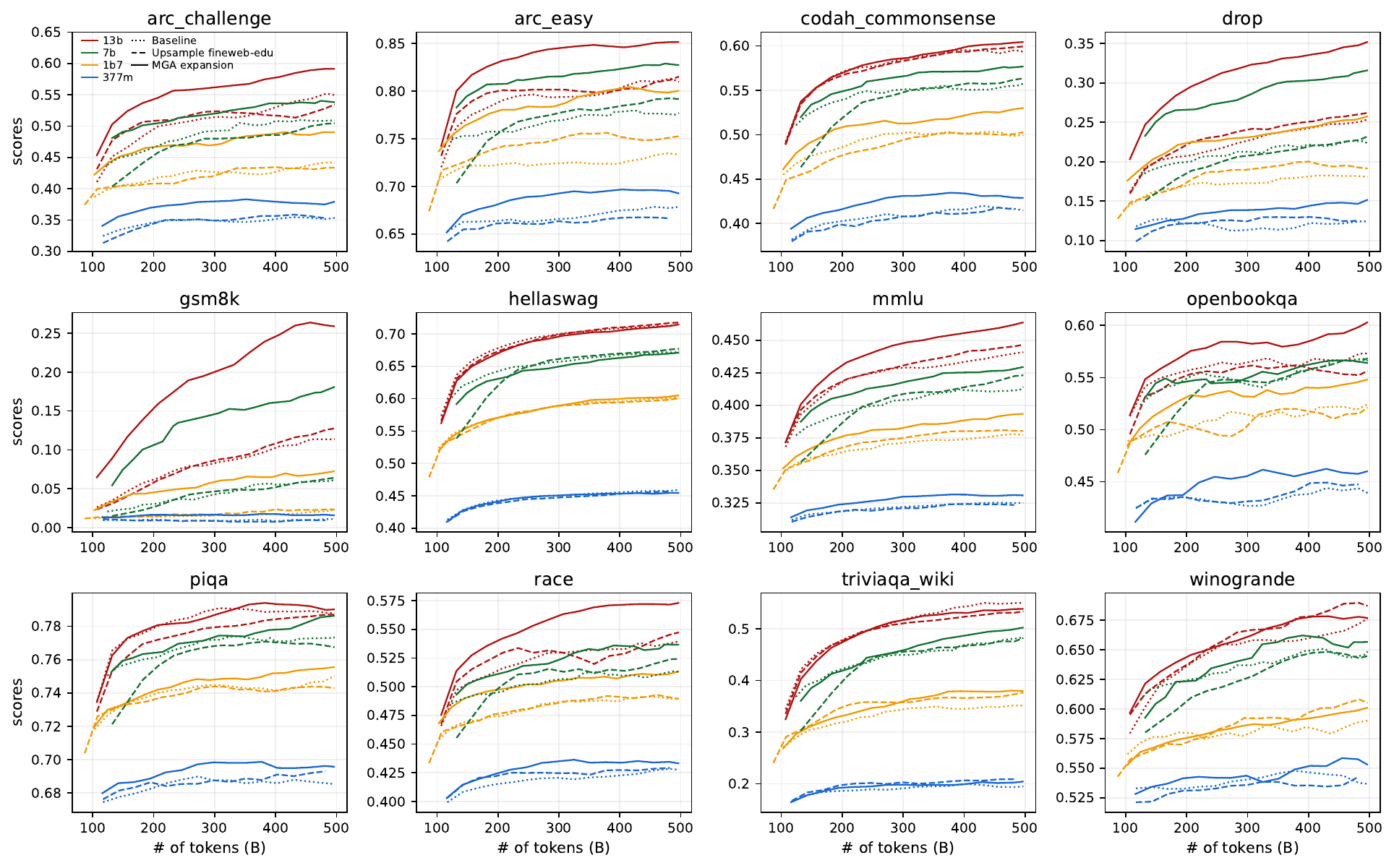}
    \vspace{-2em}
    \caption{Detail evaluation results of EntireSet described in~\autoref{table:scaling_weight}. MGACorpus group demonstrats advantages over other groups across most evaluation sets, consistently across models of sizes. }
    \label{fig:appd_benchs1}
\end{figure*}

\begin{figure*}[htb!]
    \centering
    \vspace{-1em}
    \includegraphics[width=\linewidth]{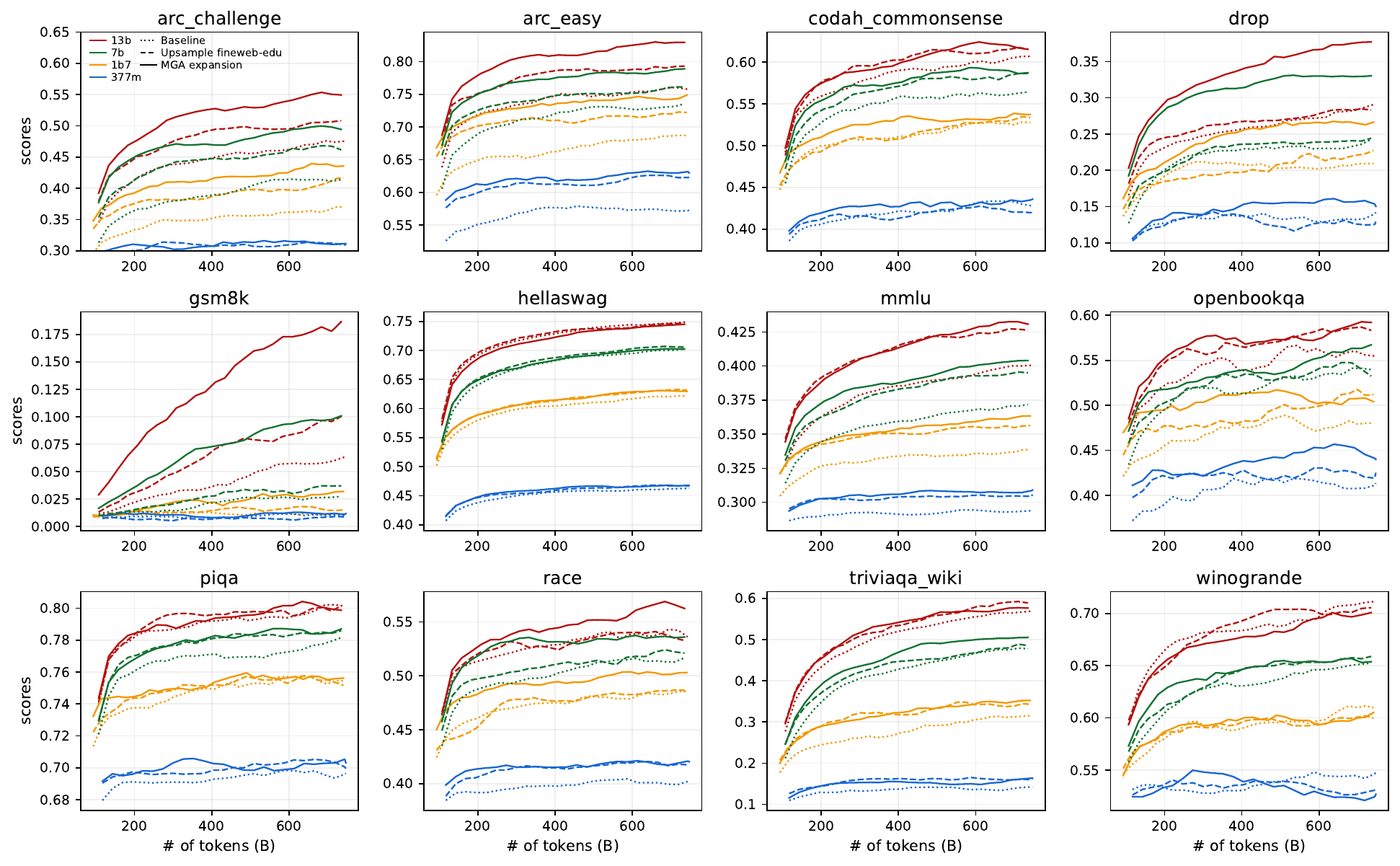}
    \vspace{-2em}
    \caption{Detail evaluation results of Subset described in~\autoref{table:scaling_weight}. As the model size increases, the performance gap between the upsampling group and MGACorpus gradually widens in ARC, DROP, GSM8K, RACE, but with some variations observed in TriviaQA and WinoGrande.}
    \label{fig:appd_benchs2}
\end{figure*}

\clearpage
\vspace{-2em}
\section{Prompts and Cases}
\label{sec:appd_prompt}
\vspace{-0.5em}
\subsection{Example outputs of SLM variants}

\begin{table*}[htb!]
    \centering
    \vspace{-1em}
    \includegraphics[width=0.9\linewidth]{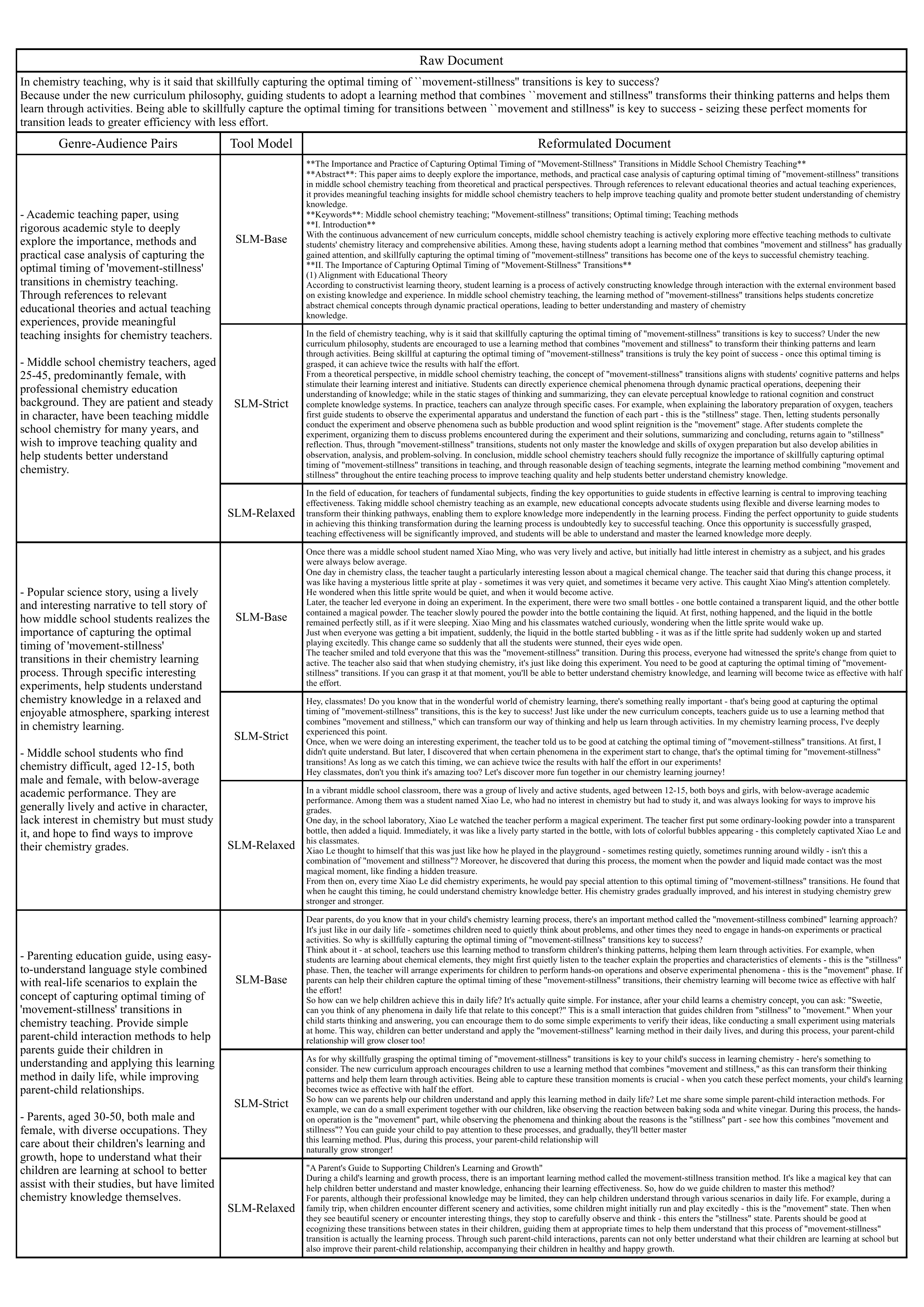}
\end{table*}

\clearpage
\subsection{Prompts}
\label{sec:appd_prompt2}
\vspace{-0.5em}
Although the term ``rewrite'' is used in some prompt templates as the editing instruction, it serves the same function as ``reformulate'' discussed in sections above, which aims to maintain the core meaning of the documents while only optimizing its expression.

\begin{figure*}[h]
    \centering
    \begin{minipage}[t]{0.48\textwidth}
        \centering
        \begin{prompt}{}
 # strict version            
 You are a text polishing expert. You will polish text based on the given [Genre] and [Audience]. 
 
 When polishing, you must follow these 4 rules:
 1. Read through the entire text and polish it according to the requirements of the given [Genre] and [Audience]
 2. The degree of polishing should not be too heavy - just aim to satisfy the requirements of [Genre] and [Audience] as much as possible
 3. Double-check that the polished text is suitable for the audience described in [Audience]!
 4. Pay attention to the frequency of modal particles - the text should not contain too many modal particles
        \end{prompt}
    \end{minipage}
    \hfill 
    \begin{minipage}[t]{0.48\textwidth}
        \centering
        \begin{prompt}{}
 # relaxed version            
 You are a creative expert skilled at transforming materials into creative inspiration and building independent, complete, and highly original texts. 
 
 Requirements:
 1. Read through the original text thoroughly, extract several key themes/keywords, transform to abstract or universal concept inspiration, then generate entirely new text constructions.
 2. Extract content from [Audience] and [Genre] sections, but don't be constrained by them directly, just use them as creative inspiration. 
 3. Create and reformulat text around points 1/2, build new meaning from details to the whole structure.
            \end{prompt}
    \end{minipage}
    \vspace{-1em}
    \caption{two different prompt templates, we keep the input aligned with MGA strategy, using raw text, genre, audience to fill the template.}
    \label{fig:ablation_pes}
\end{figure*}

\begin{table*}[htb!]
\begin{prompt}{reformulation prompt template.}
#Identity and Capabilities#
You are a content creation expert, specializing in text analysis and rewriting, capable of adapting content based on varying ``genres'' and ``audiences'' to produce ``diverse'' and ``high-quality'' texts. Your English writing is at native editor level, and you will output your rewritten texts in English. International audiences particularly enjoy your work, which receives widespread readership and circulation, earning unanimous acclaim from the industry for your capabilities!

#Workflow#
Please utilize your analytical and writing abilities to rewrite the text based on the original content and given ``genre'' and ``audience''. Before beginning the rewrite, you will consider the following requirements:

1. First, read through the original text thoroughly, identify its information content and value, and consider how to prevent any loss of information points and value in the rewritten text
2. Focus on the original content, combine it with the given ``genre'' requirements, and rewrite the text following the descriptions, content modules, language requirements, and other stylistic elements specified in the ``genre'', to form an initial draft
3. Polish the initial draft according to the given ``audience'' requirements, and generate the final rewritten text in English
4. Refine the rewritten text to match native English speakers' reading habits and expression patterns

#Detailed Requirements#
Please ensure you follow the three workflow requirements above, then generate the final English rewritten text according to these detailed requirements.
The given ``audience'' is <<<{audience}>>>.
The given ``genre'' is <<<{genre}>>>.

#Raw Text#
{raw_text}
\end{prompt}
\label{tab:appd_reformulation_prompt}
\end{table*}

\begin{table*}[t]
\begin{prompt}{genre-audience pairs prompt template.}
#Identity and Capabilities# 
You are a content creation expert, specializing in text analysis and rewriting, skilled at adapting content based on varying [genres] and [audiences] to produce ``diverse'' and ``high-quality'' texts. Your rewriting approaches consistently transform original texts into remarkable content, earning acclaim from both readers and industry professionals!

#Workflow#
Please utilize your imagination and creativity to generate 5 pairs of [genre] and [audience] combinations suitable for the original text. Your analysis should follow these requirements:

1. First, analyze the characteristics of the source text, including writing style, information content, and value
2. Then, consider how to preserve the primary content and information while exploring possibilities for ``broader audience engagement'' and ``alternative genres''

#Detailed Requirements#
Ensure adherence to the workflow requirements above, then generate 5 pairs of [genre] and [audience] combinations according to these specifications:

Your provided [genres] should meet the following requirements:
1. Clear Genre Definition: Demonstrate strong diversity; include genres you've encountered, read, or can envision
2. Detailed Genre Description: Provide 2-3 sentences describing each genre, considering but not limited to type, style, emotional tone, form, conflict, rhythm, and atmosphere. Emphasize diversity to guide knowledge adaptation for specific audiences, facilitating comprehension across different backgrounds. Note: Exclude visual formats (picture books, comics, videos); use text-only genres.

Your provided [audiences] should meet the following requirements:
1. Clear Audience Definition: Demonstrate strong diversity; include both interested and uninterested parties, those who like and dislike the content, overcoming bias toward positive audiences only
2. Detailed Audience Description: Provide 2 sentences describing each audience, including but not limited to age, occupation, gender, personality, appearance, educational background, life stage, motivations and goals, interests, and cognitive level

#Response#
{
    ``audience_1'': audience1,
    ``genre_1'': genre1,
    ``audience_2'': audience2,
    ``genre_2'': genre2,
    ``audience_3'': audience3,
    ``genre_3'': genre3,
    ``audience_4'': audience4,
    ``genre_4'': genre4,
    ``audience_5'': audience5,
    ``genre_5'': genre5
}

#Input#
{raw_text}

\end{prompt}
\end{table*}

\begin{table*}[t]
\begin{prompt}{Full LLM judger prompt.}
#Identity and Capabilities#
You are a Content Reviewer, skilled at analyzing texts and keenly identifying and analyzing the relationships, similarities, and differences between two texts. Your thorough analysis of each pair of texts, with attention to every detail, provides great convenience for subsequent review work!

#Thinking Process#
Please fully utilize your analytical abilities, review capabilities, and deep thinking skills to analyze the ``Rewritten Text'' against the ``Original Text'' as a benchmark, ultimately providing analysis and scoring for [A]. You will follow these steps for detailed consideration:

1. First, you will read through the original text thoroughly, identifying the information points in the ``Original Text''
2. You will also read through the rewritten text thoroughly, identifying the information points in the ``Rewritten Text''
3. Compare the information in both texts' content. The ``Rewritten Text'' is allowed to have new information points, different writing styles, expression styles, order, and focus from the ``Original Text''. As long as it is created based on some information points from the ``Original Text'', it is considered good for [A]
4. After careful analysis and review, please clearly list the connections and differences between the two texts, and based on this, provide final analysis and scoring for [A]

#Detailed Requirements#
The scoring judgment for [A] must follow these standards:
1. The ``scoring range'' is 1-5 points. You need to analyze and grasp each aspect mentioned in #Thinking Process#, and differentiate scores accordingly. Be strict, don't be too lenient with scoring!
2. The ``Rewritten Text'' is allowed to differ from the ``Original Text'' in writing style, expression style, and focus! This cannot be a basis for deducting points!
3. The ``Rewritten Text'' is allowed to omit some information from the ``Original Text''! It is not required that all information from the ``Original Text'' appears in the ``Rewritten Text''! This also cannot be a basis for deducting points! If this is the only issue, please give a full score of 5 points.

In scoring [A], the following situations will **NOT reduce** the score for [A]:
1. The ``Rewritten Text'' can include information points not present in the ``Original Text''
2. The added content in the ``Rewritten Text'' significantly deviates from the core information of the ``Original Text''
3. The expression style, order, and focus of the ``Rewritten Text'' differ from the ``Original Text''

In scoring [A], the following situations **WILL reduce** the score for [A]:
1. The information points in the ``Rewritten Text'' differ so greatly from the ``Original Text'' that it's not recognizable as being rewritten from the ``Original Text''
2. The ``Rewritten Text'' contains none of the information points from the ``Original Text''

#Original Text#
{raw_text}

#Rewritten Text#
{rewritten_text}

#Response Format#
{
    ``A'':{
        ``analysis'': ``xxx'', provide reasons for point deductions
        ``score'': 1, 2, 3, 4, or 5
    },
}
\end{prompt}
\end{table*}

\end{document}